%% file: main.tex
\documentclass[a4paper,journal]{IEEEtran}
\usepackage{amsmath,amsfonts}
\usepackage{algorithmic}
\usepackage{algorithm}
\usepackage{array}
\usepackage{pifont}
\newcommand{\xmark}{\ding{55}}  
\usepackage{subcaption}  
\usepackage[table]{xcolor}
\definecolor{lightgreen}{rgb}{0.88, 1, 0.88}  
\usepackage{textcomp}
\usepackage{stfloats}
\usepackage{url}
\usepackage{verbatim}
\usepackage{multirow}
\usepackage[normalem]{ulem}
\useunder{\uline}{\ul}{}
\usepackage{graphicx}
\usepackage{cite}
\usepackage[export]{adjustbox}  
\usepackage{booktabs}
\usepackage{float}
\usepackage{wrapfig}  
\usepackage{hyperref}
\usepackage{cleveref}

\begin{document}

\title{Event-to-Video Reconstruction using Spatio-Temporal and Frequency-Enhanced Deep Neural Networks \\

\thanks{}
}
\author{
    Ramna Maqsood, Paulo Nunes, Luís Ducla Soares, and Caroline Conti \\
    \textit{Instituto de Telecomunicações, Instituto Universitário de Lisboa (ISCTE-IUL)} 
    Lisbon, Portugal \\
    
}

\maketitle

\begin{abstract}
Event cameras offer significant advantages over conventional frame-based counterparts, including high temporal resolution, low latency, and energy efficiency. These characteristics make them suitable for high-speed and high-dynamic range scene acquisition scenarios; however, the lack of dense intensity frames limits the direct applicability of conventional computer vision methods for scene understanding. Event-to-video (E2V) reconstruction seeks to bridge this gap by converting asynchronous event streams into a sequence of synchronous video frames. Existing E2V reconstruction methods based on convolutional neural networks (CNNs) and transformer models operate primarily in the spatial domain and often struggle to recover fine structural details while suppressing severe reconstruction artifacts. To address these issues, we propose MSFET-E2V, a novel multiscale frequency-enhanced transformer model. At its core lies a cross-domain attention module (CDAM), which fuses spatio-temporal features with frequency-aware representations derived from the discrete wavelet transform (DWT). Unlike prior methods relying solely on spatial attention, our approach effectively captures both local and global structures by taking into account low- and high-frequency components, enhancing detail preservation and robustness across various motion scenarios. Furthermore, we propose a lightweight wavelet-enhanced skip block (WSB) that serves as a skip connection, facilitating artifact suppression and structural detail refinement through joint spatial–frequency domain processing. Extensive experiments demonstrate that MSFET-E2V achieves superior performance over state-of-the-art methods on multiple real-world event datasets, offering significant gains in reconstruction quality. Moreover, compared to the existing transformer-based method, our proposed model significantly reduces the number of parameters, the GPU memory usage, and inference time for the E2V reconstruction task.  
\end{abstract}


\section{Introduction}
Event cameras are bio-inspired vision sensors that operate asynchronously, recording intensity changes at the pixel level rather than capturing full-frame images at fixed time intervals. This unique property enables event cameras to achieve high temporal resolution and low scene acquisition latency, making them well-suited for scenarios involving fast motion and challenging lighting conditions. In recent years, event-based vision has gained significant attention in the research community due to its potential applications in robotics\cite{robot}, autonomous driving\cite{auto_driv} and surveillance\cite{aug_reality}. One of the key challenges in event-based processing is the absence of conventional image frames, which has led researchers to explore event-to-video (E2V) reconstruction, a critical task aimed at converting sparse event streams into dense, frame-based representations. High-quality video reconstruction from events enables better visualization, facilitates compatibility with traditional vision algorithms, and enhances downstream tasks like object recognition and tracking\cite{obj_track}.

A significant line of research in E2V reconstruction has focused on convolutional neural network (CNN) architectures\cite{E2VID,E2VID+,SPADE-E2VID}, which have demonstrated strong feature extraction capabilities. Unlike conventional video frames, event streams encode only localized intensity changes over time, making it difficult for CNNs, designed with local receptive fields, to effectively model long-range dependencies and global correlations in the data. This limitation is particularly pronounced in regions with low event density, where CNNs struggle to reconstruct structural details. One commonly explored approach to mitigate this issue has been to increase the convolutional kernel size to expand the receptive field. However, empirical evidence suggests that beyond a certain point, increasing the kernel size does not yield further improvements in reconstructing structural scene details and may even degrade performance due to excessive parameterization and the loss of these structural details\cite{WaveConv,ELWNet}.

To address these limitations, recent efforts have turned toward transformer neural networks, which have been shown to achieve state-of-the-art (SOTA) results for E2V reconstruction \cite{ET-Net, transformer_2}. The main success behind transformers is their ability to model long-range contextual dependencies by applying a grid-based self-attention mechanism on image patches (i.e., tokens). However, the existing transformer-based models\cite{ET-Net,transformer_2} suffer from two limitations: (i) the quadratic scaling of the existing models with input size imposes a significant computational burden, which contradicts the low-latency nature of event cameras; (ii) another key challenge associated with the transformer-based methods is that self-attention mechanisms, while effective at capturing global dependencies, can struggle to preserve fine-grained details; this is because they tend to focus on larger, more prominent features while potentially overlooking subtle variations or textures\cite{fet}. Consequently, the self-attention mechanism alone may not fully exploit the fine-grained structures present in event streams, resulting in the loss of details in reconstructed scenes and the appearance of blurry or ghosting artifacts.

In summary, existing E2V reconstruction models suffer from three primary limitations: (i) inadequate modeling of long-range dependencies by CNN-based models due to limited receptive fields; (ii) high computational overhead and memory usage in transformer-based models, particularly for high-resolution inputs; and (iii) insufficient recovery of fine-grained structures and suppression of artifacts due to the lack of frequency domain awareness in conventional attention mechanisms.

Recently, frequency-based methods have gained attention in E2V reconstruction for their ability to effectively capture multiscale structures and mitigate the limitations of pure spatial domain approaches by employing the inherent sparsity and high temporal resolution of event data\cite{deep_unfold_TIP,spike_fft}. In \cite{trans_fft}, the authors deploy a fourier transform (FT)-based transformer model to operate entirely in the frequency domain for event-driven tasks. FT decomposes signals into their constituent sinusoidal components, providing a global frequency representation that can complement the spatial-domain limitations of the self-attention mechanism. One key limitation of FT in the context of event-based vision is its global nature. FT analyzes the input globally, transforming it into a frequency spectrum without preserving localized spatial structures. We argue that time-frequency decomposition can serve as a more effective and efficient representation space for event-based vision, considering their sparse and robust properties\cite{WaveCNet}. Unlike FT, time-frequency methods such as the discrete wavelet transform (DWT) provide a localized frequency analysis, allowing the model to capture low-frequency (LF) and high-frequency (HF) components at multiple scales and in specific regions of the signal. In the context of DWT, for 1D signals, like audio, `time' refers to the temporal location in the sampling dimension, while for 2D signals, such as images, it corresponds to the spatial location, enabling analysis of where specific spatial frequency components occur. In fact, a study\cite{wave_hm} on human visual perception suggests that the human visual system discerns elementary features through time-frequency components rather than purely global frequency representations. This biological inspiration further supports the idea that event-based vision, which mimics certain aspects of human perception, can benefit from time-frequency representation.

In this work, we investigate the use of event data coupled with deep neural networks for the E2V reconstruction task, operating in three domains: (i) spatial, (ii) temporal and (iii) frequency domains. Our primary objective is to address the limitations of existing CNN and transformer-based models for the considered task. To achieve this, we propose a multiscale frequency-enhanced transformer model (MSFET-E2V) for E2V reconstruction. The event stream encodes sparse intensity changes over time, which implicitly contain both HF details and LF structural cues. To effectively exploit these characteristics of event data, we propose a cross-domain attention module (CDAM). This module integrates spatio-temporal features, extracted via a convolutional recurrent neural network (CRNN), with frequency domain representations obtained through DWT. We use DWT to decompose input features into multi-resolution subbands, allowing the model to separately process LF structures and HF details, where most event-based cues reside — thus enabling more effective cross-domain fusion and a larger receptive field \cite{WaveConv} with minimal computational cost. Additionally, motivated by\cite{Day-Night}, the observation that HF components in event data can resemble noise-like patterns, we introduce a wavelet-enhanced skip block (WSB) designed to refine intermediate features and suppress artifacts. Inspired by these recent findings in image restoration, our module helps disambiguate signal from event-induced noise, enhancing the clarity and realism of the reconstructed images. Finally, we propose a residual-guided decoder (RGD) module responsible for improving the fidelity of the reconstructed intensity images. Our contributions are as follows:
\begin{itemize}
    \item In this work, we propose MSFET-E2V, a novel multiscale frequency-enhanced transformer model designed for high-quality E2V reconstruction. Our model introduces three key innovations: (i) a CDAM that effectively fuses spatio-temporal features with frequency-aware representations, enabling improved structural detail preservation of the scene, (ii) WSB that serves as a skip connection, refining fine details and suppressing reconstruction artifacts through joint spatial-frequency domain processing and (iii) RGD module that employs residual layers to preserve the features and for stable gradient propagation during reconstruction.
    \item Validation via extensive experiments on various event-based datasets shows that this model can generate higher-quality video reconstruction than the previous SOTA methods in various scenarios. The MSFET-E2V improves efficiency, achieving over 70\% faster inference speed and reducing GPU memory usage by more than 50\% compared to the existing transformer-based video reconstruction methods. 
\end{itemize}

\section{Related Work}

\subsection{Event-based Video Reconstruction}
In the realm of E2V reconstruction, deep architectures have demonstrated remarkable efficacy. Rebecq \textit{et al}\cite{E2VID} were the pioneers in introducing a novel CNN-based model called E2VID based on the encoder-decoder architecture. In their work, the authors managed to demonstrate the superiority of a learning-based approach over handcrafted methods\cite{3dTracking}. Scheerlinck \textit{et al} \cite{FireNet} put forth a lightweight model named “FireNet”, which is a simplified version of E2VID. FireNet offers lower memory consumption and faster inference with a compromise in reconstruction quality. Stoffregen \textit{et al}\cite {E2VID+} introduced the E2VID+ and FireNet+ models, further enhancing results by using a novel approach based on statistics from synthetic datasets to match real-world test data. Cadena \textit{et al} \cite{SPADE-E2VID} proposed SPADE-E2VID model by incorporating a new decoding layer in the E2VID model known as the “spatially adaptive denormalization layer” and improved the reconstruction quality in early frames. Recently, the HyperE2VID \cite{HyperE2VID} model was proposed, showcasing a dynamic neural network that incorporates a context fusion module with hypernetworks\cite{hypernetworks}, which effectively reduces both computational requirements and improves reconstruction results. More recently, the work in \cite{Temporal-Diff} introduced a diffusion-based model that utilizes three conditioning signals: temporal features, LF textures derived from intensity images that represent the coarse structural information of the scene, and HF events, which correspond to fine details such as edges and rapid motion.
To the best of our knowledge, the works that incorporated transformers in E2V reconstruction were proposed by Weng \textit{et al}\cite{ET-Net} and Gu \textit{et al}\cite{transformer_2}, aiming to better exploit the global context of input data. However, their complex architectures resulted in a loss of details in the intensity image during reconstruction due to their low-pass self-attention mechanism and, additionally, it incurred higher inference times. Despite the promising potential of transformers in event-based systems, their utilization in E2V reconstruction, especially in conjunction with frequency domain analysis, has not yet been explored. Consequently, our approach of combining transformers with wavelet-based frequency domain analysis is novel and aims to address this gap, aiming to preserve scene details in the reconstructed intensity images.

\subsection{Deep Discrete Wavelet Transform Models}
Recently, studies on deep frequency-based analysis have attracted considerable attention in the field of event-based vision. In \cite{deep_unfold_TIP}, the authors proposed to utilize spatial-frequency domain knowledge for E2V reconstruction to overcome the limitations of existing methods\cite{E2VID, ET-Net, HyperE2VID} that only rely on spatial domain by incorporating a fourier-based regularization term. Another work\cite{spike_fft} proposed a temporal-frequency calibration module based on the FT to improve the contrast of the E2V reconstructions. However, both methods rely on FT for analyzing frequency components, which, while effective for stationary signals, struggles with localized and non-stationary event-based data due to its lack of time-frequency localization. The DWT, on the other hand, can better handle transient non-stationary signals through its multi-resolution capability\cite{3D-DWT,SWFormer}. Building on these strengths, Yao \textit{et al}\cite{wave-vit} proposed Wave-ViT, a novel vision transformer that uses wavelet-based downsampling to address the limitations of traditional CNN-based pooling and stride methods. Unlike traditional approaches that cause irreversible information loss, the Wave-ViT integrates invertible wavelet transforms to preserve fine and structural details while maintaining computational efficiency for various vision tasks. Recently, Fang \textit{et al}\cite{SWFormer} proposed a spike wavelet transform (SWformer) model, which integrates DWT into spiking neural networks (SNNs). Their work demonstrates that SWformer can efficiently and effectively extract both spatial and frequency features from event data, outperforming conventional SNNs in various vision tasks. Ieng \textit{et al} \cite{Linear_basis} explores the computational complexity and efficiency of iterative basis transformations, such as DWT, when applied to event-based signals. The authors analyze how iterative transformations can effectively capture the sparse and dynamic nature of event data while maintaining computational efficiency. Their work demonstrates that such transformations not only reduce the computational burden but also enhance the quality of feature extraction, making them well-suited for real-time event-based vision applications.
\begin{figure}[t]
    \centering
    \includegraphics[width=3.2in]{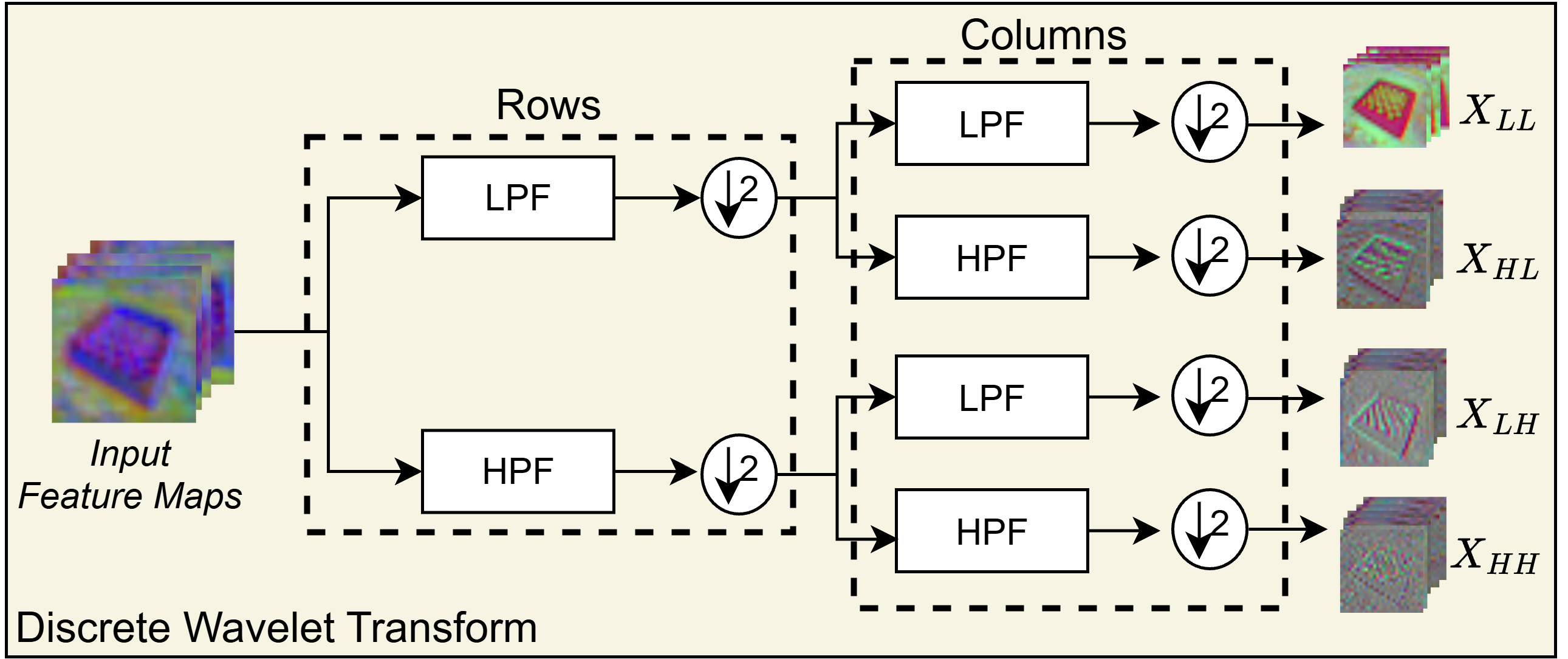} 
    \caption{Workflow of DWT, applying low-pass filter (LPF) and high-pass filter (HPF) along rows and columns separately to each channel of the feature map.}
    \label{fig:fig1}
\vspace{-4mm}
\end{figure}

Despite the growing interest in frequency-based analysis and its integration with deep learning, no prior work has fully explored the potential of DWT with a deep learning model within the domain of E2V reconstruction.
\begin{figure*}[t]
    \centering
    \includegraphics[width=0.95\linewidth]{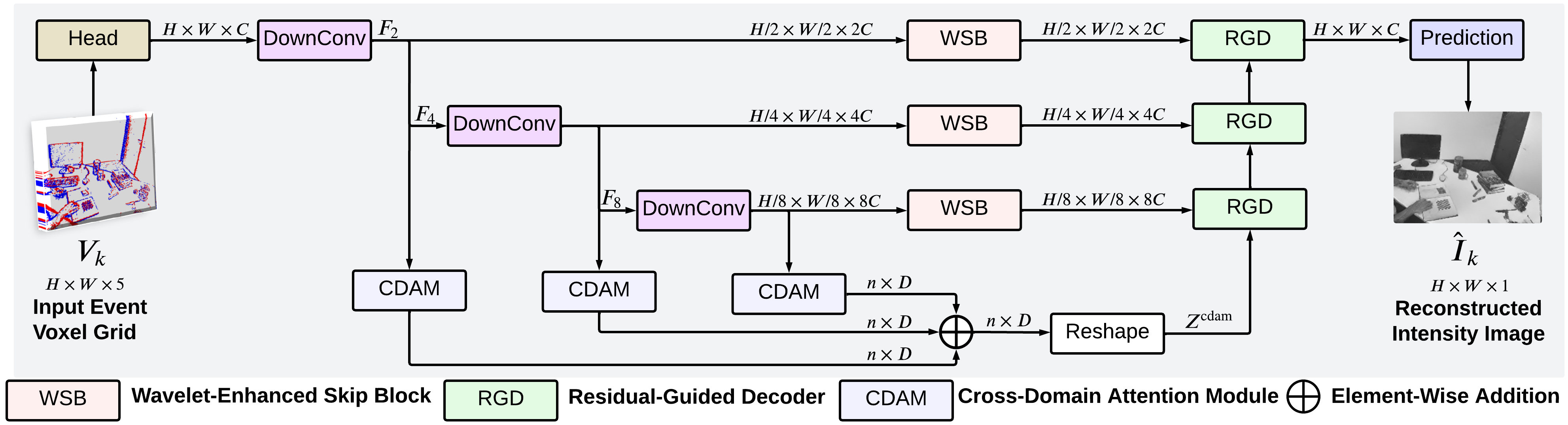}
    \caption{Overall architecture of the proposed MSFET-E2V model. DownConv blocks transforms the input event features into multiscale representations.   CDAM performs a dual-domain fusion strategy based on spatio-temporal and frequency pathways to extracts deep local and global features from each scale. WSBs bridge the encoder and decoder by refining intermediate features in both spatial and frequency domains. Finally, the reconstruction of grayscale image is performed through a series of RGD blocks.}
    \label{fig:fig2}
\vspace{-4mm}
\end{figure*}
\section{Proposed Methodology}
This paper considers the problem of reconstructing a sequence of intensity images, \( \{ \hat{I}_k \} \), from an event stream. Starting from the common approach in the literature\cite{voxel,HyperE2VID}, the event stream is converted into voxel grids. Each voxel grid is then passed to our proposed model which reconstructs a 2D grayscale image from it, having dimensions $H \times W \times 1$, where $H$ and $W$ represent the image height and width, respectively.

\subsection{DWT Definition}
Before presenting our proposed methodology, we first briefly describe the working of DWT for frequency domain analysis. DWT decomposes signals into multiscale frequency components, allowing localized analysis of both LF and HF information, which is particularly effective for image decomposition. In contrast to CNNs, which rely on localized convolutional filters and require stacking of multiple layers or enlargement of kernel sizes to increase the receptive field, DWT inherently enables hierarchical feature extraction through spatial downsampling and frequency decomposition. This leads to a more efficient expansion of the receptive field without incurring additional computational overhead or parameter count\cite{dwt_dblur, WaveCNet}. In the context of 2D input feature maps, the DWT decomposes each input feature map, \( \mathbf{X} \in \mathbb{R}^{H \times W} \), into four subbands, each representing different frequency components: low-low (LL), low-high (LH), high-low (HL), and high-high (HH).  For example, using the Haar wavelet\cite{haar}, one of the simplest and most efficient wavelets, the transformation applies a low-pass filter (LPF) and a high-pass filter (HPF) to the rows and columns of the image sequentially (as shown in Fig. \ref{fig:fig1}). Formally, the four convolutional filters for Haar are defined in equation (\ref{eq:haar_filters}):
\begin{align}
f_{LL} &= \frac{1}{2} 
\begin{bmatrix}
1 & 1 \\
1 & 1 
\end{bmatrix}, \quad
&f_{LH} &= \frac{1}{2} 
\begin{bmatrix}
-1 & -1 \\
1 & 1 
\end{bmatrix}, \nonumber \\
f_{HL} &= \frac{1}{2} 
\begin{bmatrix}
-1 & 1 \\
-1 & 1 
\end{bmatrix}, \quad
&f_{HH} &= \frac{1}{2} 
\begin{bmatrix}
1 & -1 \\
-1 & 1 
\end{bmatrix}
\label{eq:haar_filters}
\end{align}
The LL filter, \( f_{LL} \), captures LF information, while \( f_{LH} \), \( f_{HL} \), and \( f_{HH} \) are designed to extract HF information in three distinct directions: horizontal, vertical, and diagonal, respectively. More specifically, a one-dimensional DWT is applied by convolving LPF and HPF along the row and column directions of the input separately. This results in four subband images — \( X_{LL} \), \( X_{LH} \), \( X_{HL} \), and \( X_{HH} \), which together retain all essential image information. According to this formulation, the decomposition and convolution process is performed using equation (\ref{eq:decompose}):
\begin{equation}
\begin{aligned}
X_{LL} &= (f_{LL} * X) \downarrow 2, \quad &X_{LH} &= (f_{LH} * X) \downarrow 2, \\
X_{HL} &= (f_{HL} * X) \downarrow 2, \quad &X_{HH} &= (f_{HH} * X) \downarrow 2,
\end{aligned}
\label{eq:decompose}
\end{equation}
where \( * \) denotes the 2D convolution operation and \( \downarrow2 \) represents downsampling by a factor of 2. The key advantage of DWT over CNNs lies in its capacity to capture both fine-grained and long-range spatial correlations in a single transformation\cite{wave-vit}. In CNNs, the receptive field grows gradually by stacking layers, which may still be insufficient to capture large-scale dependencies unless deeper architectures or larger kernels are used, both of which come at the cost of increased computational burden and risk of over-smoothing\cite{WaveConv}. DWT, on the other hand, effectively expands the effective receptive field through its hierarchical decomposition. 
\subsection{Event Representation}
In this section, we describe our methodology for representing event data as a voxel grid. Given an event stream \(\{ e_i \}\) containing \(N_E\) events, each event \(e_i = (x_i, y_i, t_i, p_i)\) corresponds to the \(i\)-th event characterized by spatial coordinates \((x_i, y_i)\), timestamp \(t_i \in [0, T]\), and polarity \(p_i \in \{+1, -1\}\), indicating an increase $(+1)$ or decrease $(-1)$ in brightness, respectively, for all \(i \in \{0, \ldots, N_E - 1\}\). Assuming that the ground truth (GT) intensity frames are available with the incoming event stream, one can group events such that every event between consecutive frames ends up in the same group. Specifically, given the frame timestamps, the set of events in the \(k\)-th group is defined as $E_k = \{ e_i \mid T_{k-1} < t_i \leq T_k \}$, where \(T_k\) is the ending timestamp of the \(k\)-th group and \(\Delta T = T_k - T_{k-1}\) is its duration. We divide the events in each group into $B$ temporal bins to discretize the continuous-time event data; in the current work we adopted $B=5$. The timestamps of the events in the group are first normalized to the range \([0, B-1]\). Each event then contributes its polarity to the two closest bins using bilinear interpolation. The voxel grid \(V_k \in \mathbb{R}^{H \times W \times B}\) for the \(k\)-th group is formed using equation (\ref{eq:voxel_grid}):
\begin{equation}
V_k(x, y, t) = \sum_i p_i \max(0, 1 - |t - t^*_i|)  \delta(x - x_i, y - y_i)
\label{eq:voxel_grid}
\end{equation}
where $\delta$ is the Kronecker delta that is used to select the event location and $t_i^*$ is the normalized timestamp calculated using equation (\ref{eq:timestamp}):
\begin{equation}
t^*_i = (B - 1) \frac{t_i - T_k}{\Delta T}
\label{eq:timestamp}
\vspace{-4mm}
\end{equation}
\begin{figure*}[t] 
    \centering
   \includegraphics[width=\linewidth]{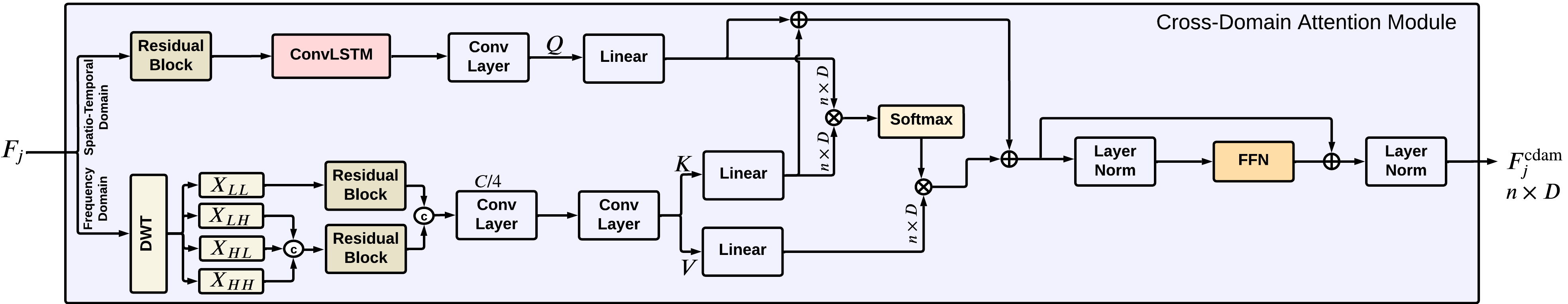} 
    \caption{Detailed structure of the proposed cross-domain attention module (CDAM). The module employs a cross-domain multi-head self-attention mechanism, where $Q$, $K$, and $V$ are derived from different domains, specifically, the spatio-temporal and frequency domains, to enable complementary feature fusion.}
    \label{fig:fig3}
\end{figure*}
\begin{figure*}[t]
    \centering
    \begin{subfigure}{0.48\linewidth}
        \centering
        \includegraphics[width=\linewidth]{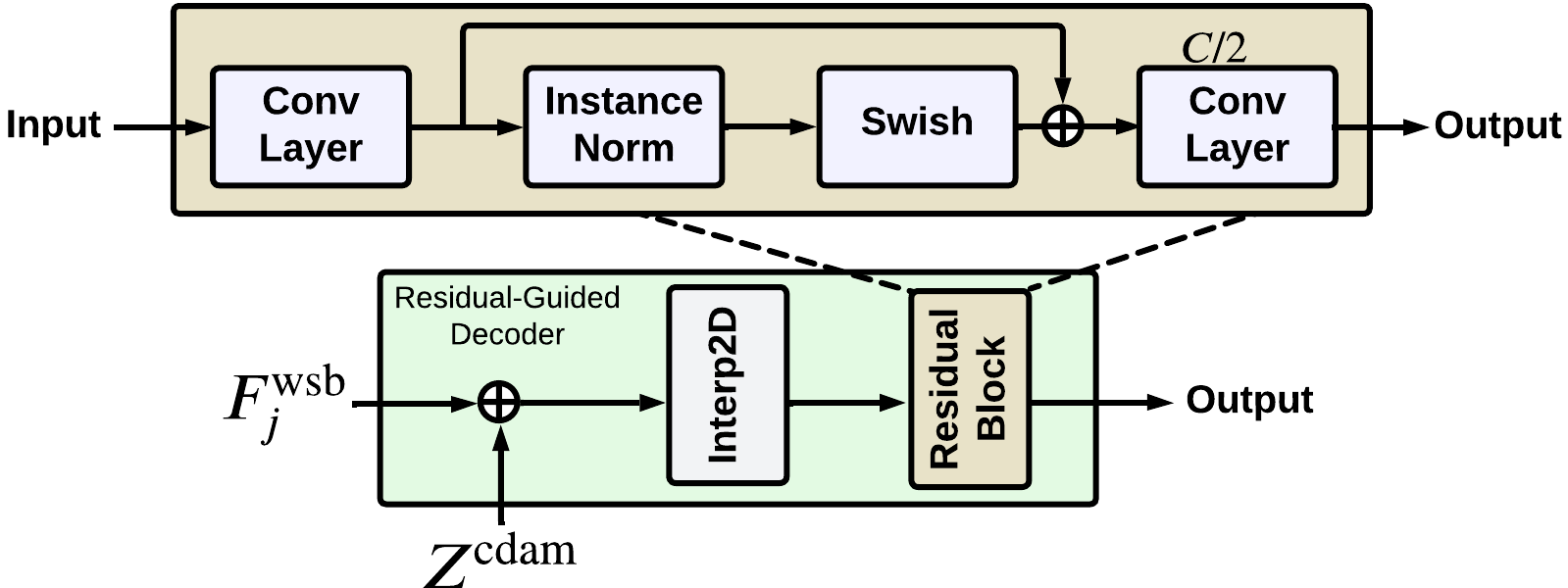}
        \caption*{(a)}
        \label{fig:rgd_block}
    \end{subfigure}
    \hfill
    \begin{subfigure}{0.48\linewidth}
        \centering
        \includegraphics[width=\linewidth]{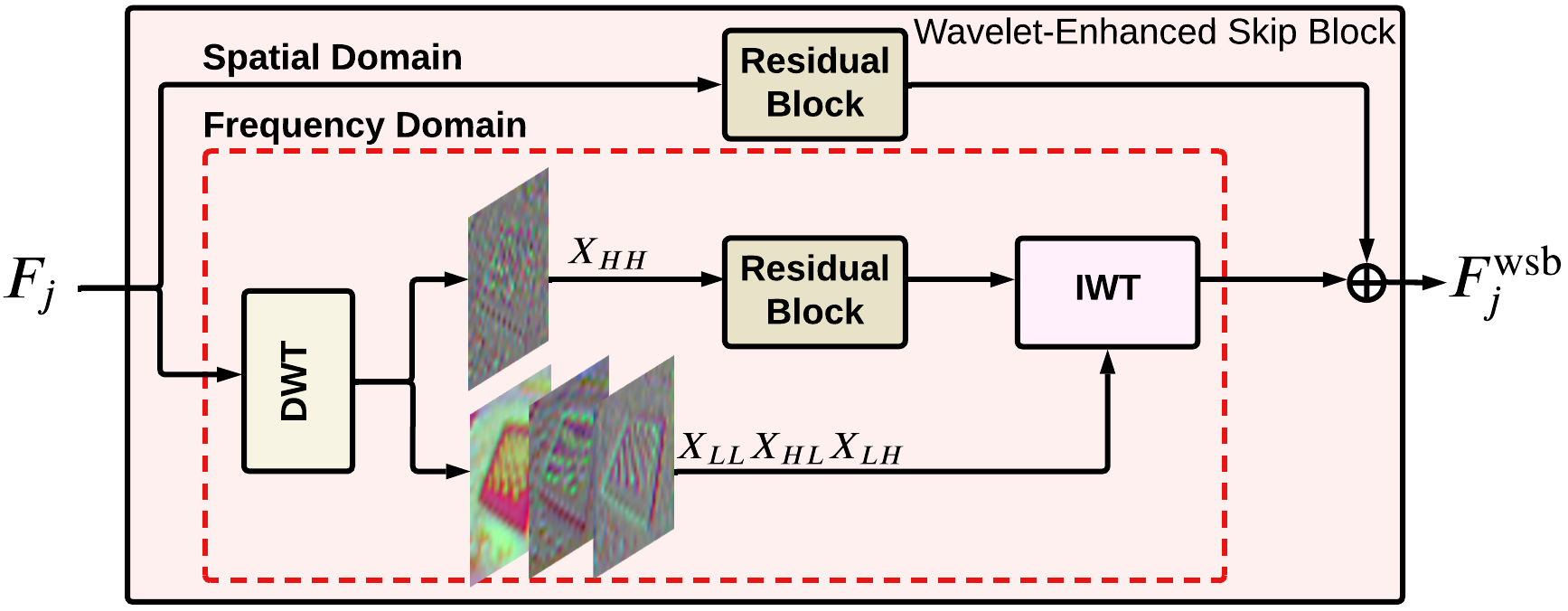}
        \caption*{(b)}
        \label{fig:wsb_block}
    \end{subfigure}
    \caption{Detailed internal structure of the RGD and WSB: (a) The diagram illustrates the working of the first RGD block, where the input is taken from both the WSB, $F_j^{\mathrm{wsb}}$, and the fused CDAM features, $\mathrm{Z}^{\mathrm{cdam}}$. The remaining two RGD blocks share the same internal structure but differ in that their inputs. (b) The proposed WSB is responsible for fusing shallow features from DownConv blocks with deep features from RGD blocks to reduce artifacts and enhance feature consistency.}
    \label{fig:fig4}
\vspace{-4mm}
\end{figure*}
\subsection{Proposed MSFET-E2V Architecture}
After representing each event group as a voxel grid, our task is to generate a corresponding image stream from the sequence of voxel grids. Fig.~\ref{fig:fig2} illustrates the overall architecture of the proposed MSFET-E2V model. We adopt a U-Net-based \cite{unet} encoder-decoder framework, where the input event voxel grid $V_k$ is processed through a series of specialized blocks to reconstruct the output grayscale image $\hat{I}_k \in \mathbb{R}^{H \times W \times 1}$. Our model performs multiscale feature aggregation in the encoder, where the DownConv blocks progressively reduce the spatial resolution and extract hierarchical features at multiple scales. These features are then refined using our proposed CDAMs, which are designed to enhance both fine and structural details. The decoder comprises RGD blocks that receive enriched features from the CDAMs and reconstruct the final image. In addition, WSBs act as skip connections between corresponding encoder and decoder blocks. Unlike standard skip connections, WSBs process shallow encoder features in both spatial and frequency domains, enabling effective artifact suppression and detail preservation. We describe each component of our model in more detail below:

\textbf{Head Layer:} This layer is composed of a single convolutional (Conv) layer with a $3\times3$ kernel size and stride 1, to preserve the spatial dimensions. The input event voxel grid is transformed into a higher-dimensional feature representation $H \times W \times C$, where $C$ is the base number of channels (set to 32 in our implementation). 

\textbf{DownConv:} Our model consists of three DownConv blocks that act as encoder. Each DownConv block reduces the spatial resolution of the feature maps while simultaneously increasing their channel dimensionality. Specifically, each DownConv block consists of a Conv layer with a $3 \times 3$ kernel size and stride of 2, followed by a leaky rectified linear unit (LeakyReLU) activation function. Through the DownConv blocks,we finally obtain multiscale features $\{F_j\mid j \in \{2,4,8\} \}$, which are passed to CDAM and WSB blocks subsequently.

\textbf{Cross-Domain Attention Module (CDAM):} One of the persistent challenges in E2V reconstruction lies in accurately recovering fine-grained spatial details while suppressing structural artifacts introduced during the encoding process. Transformer-based architectures like ET-Net~\cite{ET-Net} have demonstrated notable capabilities in modeling long-range dependencies through self-attention mechanism. However, their attention mechanism typically operates solely in the spatial domain, where queries ($Q$), keys ($K$), and values ($V$) are derived from the same spatial features. This design neglects frequency domain cues that are crucial for capturing sharp edges, texture discontinuities, and HF variations inherent to event data. Consequently, existing methods~\cite{E2VID, cista-lstc, HyperE2VID} often exhibit a loss of detail in the reconstructed scenes~\cite{Temporal-Diff}.

To address this, we propose CDAM that introduces a hybrid attention formulation. Specifically, CDAM incorporates frequency-aware $K$ and $V$ extracted via a DWT while retaining spatially grounded $Q$, enabling the model to aggregate complementary features across both domains. As shown in Fig.~\ref{fig:fig2}, our model employs three CDAMs, each processing one of the multiscale feature maps  $F_j \in \mathbb{R}^{\frac{H}{j} \times \frac{W}{j} \times jC}$. As illustrated in Fig.~\ref{fig:fig3}, CDAM consists of two parallel branches: (i) a spatio-temporal domain branch, and (ii) a frequency domain branch. In the spatio-temporal domain, the input feature map $F_j$ is first processes with a residual block (RB) comprising two $3 \times 3$ convolutions with swish activation function, preserving spatial structure while enriching local representation. The output is then processed by a convolutional long short-term memory module (ConvLSTM)\cite{convlstm}, which captures long short-term temporal dependencies and maintains hidden states across time. 

In parallel, motivated by \cite{wave-vit}, we utilize DWT to process input features $F_j$ to account for frequency analysis to focus on fine-grained and structural counterparts.  We utilize the Haar wavelet function \cite{haar} due to its computational efficiency to decompose each input feature map into four frequency subbands using equation (\ref{eq:decompose}): one LF subband named $X_{LL}$, and three HF subbands named $X_{LH}$, $X_{HL}$, and $X_{HH}$. Here $X_{LL}$, contains the coarse approximation and \( \{ X_{LH}, X_{HL}, X_{HH} \} \) contain the fine-grained details of the feature maps (i.e., the edges). In our CDAM, we process the subbands independently through the RBs to capture the intricate fine and structural details from the input feature maps. More specifically, the LF subband $X_{LL}$ is processed by a RB and all three HF subbands \( \{ X_{LH}, X_{HL}, X_{HH} \} \) are concatenated along channel dimensions and processed with another RB. Both processed streams are then concatenated along the channel dimension for further processing in the frequency pathway. Since the DWT operation increases the number of channels by a factor of four (e.g., an input of size \(64 \times 64 \times 64\) becomes \(32 \times 32 \times 256\) after DWT due to the decomposition of each feature map into four subbands), the final output of the frequency path must be adapted to align with the spatio-temporal path. Therefore, a \(3 \times 3\) Conv layer is employed not only to refine the merged features but also to reduce the channel dimensionality, restoring compatibility with the spatio-temporal domain in feature dimensionality. 

To prepare both spatio-temporal and frequency features for cross-domain self-attention, we apply an overlapping patch embedding strategy to convert 2D feature maps into fixed-length token sequences. At each scale, a Conv layer with carefully chosen kernel size, stride, and padding is used to project the input into a common embedding dimension \( D = 256 \). This results in a feature map of shape \( ( D, H', W') \), which is linearly transformed into \( F_j \in \mathbb{R}^{n \times D} \), where \( n = H' \times W' \). Specifically, all three scales — \( F_2 \), \( F_4 \), and \( F_8 \) — are transformed into sequences of shape \( 256 \times 256 \) (i.e., \( H' = W' = 16 \), so \( n = 256 \)). Afterwards, the representations  $Q \in \mathbb{R}^{n \times D}$ is derived from spatio-temporal domain and $K, V \in \mathbb{R}^{n \times D}$ are derived from frequency domain to perform cross-domain multi-head  self-attention mechanism using equation (\ref{eq:MHSA}):
\begin{equation}
\text{\textit{attn}} = \text{Softmax}\left( \frac{QK^{\mathrm{T}}}{\sqrt{D}} \right) V
\label{eq:MHSA}
\end{equation}
Additionally, we also introduce the residual connection to maintain the internal dependency, where frequency-enhanced features from the wavelet domain are element-wise added to the spatio-temporal domain features before being fused with the attention output. To add non-linearity and further refine the representations learned by the self-attention mechanism, the attended features are passed through a feed-forward network (FFN), followed by another residual connection and LayerNorm, producing the output \( F_j^{\mathrm{cdam}} \) at each scale.  The FFN consists of two linear transformations with a GeLU activation between them.

\textbf{Residual-Guided Decoder (RGD)}: The reconstruction phase of our model consists of three sequential RGD blocks, followed by a prediction layer. Each RGD block is designed to progressively refine the reconstruction output through spatial upsampling, feature enhancement, and residual learning. The internal structure of an RGD block includes a bilinear interpolation layer with an upsampling factor of 2, which progressively increases the spatial resolution. After upsampling, the features are refined by a RB that enhances representational capacity and helps to minimize the artifacts. 

More specifically, the input to the first RGD block comprises deep features extracted from the CDAMs. These features capture both structural and fine-grained details across multiple scales. The outputs of the three CDAMs , each with shape \( n \times D \), are first aggregated by element-wise addition and then reshaped back to three-dimensional feature \( \frac{H}{j} \times \frac{W}{j} \times D \), aligning with the expected input format of the decoder. The resulting features, denoted as \(\mathrm{Z}^\mathrm{cdam}\), are subsequently passed to the first RGD block. The remaining two RGD blocks then take as input the intermediate features produced by their preceding RGD block, thereby enabling hierarchical refinement of the reconstruction. For clarity, Fig.~\ref{fig:fig4} (a) provides a detailed view of the internal structure of a single RGD block, specifically illustrating the first RGD. While the subsequent RGD blocks follow the same architectural design, the only difference lies in the source of their inputs: the first RGD takes features from \(\mathrm{Z}^\mathrm{cdam}\) and WSB (explained later), whereas the remaining two RGD receive features solely from the previous RGD block and WSB. 
Each RB inside the RGD is composed of two Conv layers with kernel size \( 3 \times 3 \), separated by a non-linear activation function. Notably, no activation is applied after the final Conv layer to preserve the linearity of the residual connection. A residual skip connection is applied between first and second Conv layer to maintain the flow of information. Additionally, to ensure compatibility between each DownConv and its corresponding RGD block, the final Conv layer in each RGD block adjusts the channel dimensionality of the decoded feature maps to match that of the encoder features. It is important to note that the same RB structure is consistently used throughout our model. However, the final Conv layer in the RB is only configured to alter the number of output channels within the RGD blocks, where dimensional alignment with the encoder features is required. In all other instances, the RB preserves the input channel dimensionality.
\vspace{-0.1mm}

\textbf{Wavelet-Enhanced Skip Block (WSB):} We observe that directly injecting the shallow features from DownConv blocks into the RGD pipeline introduces artifacts commonly referred to as "ghosting and blur effects" in the reconstructed images (also demonstrated in ~\cref{sec:ablation}). To address this issue, we propose WSB, which serves as a refined skip connection between the DownConv outputs and the RGD blocks. This component is designed to process the shallow features before fusion, in both the spatial and frequency domains, as illustrated in Fig. \ref{fig:fig4} (b). The spatial branch processes features using a RB, while the frequency branch decomposes the input feature maps using the DWT into four subbands \( \{ X_{LL}, X_{LH}, X_{HL}, X_{HH} \} \). Given the inherent noise in event-based data, which often manifests in shallow features, we aim to suppress such noise during this phase. As noted in prior works \cite{scse, Day-Night}, event noise tends to dominate the HF components, particularly the \( X_{HH} \) subband. Inspired by this observation, we apply a RB selectively to the \( X_{HH} \) subband to enhance noise awareness, while preserving the low- and mid-frequency content \( \{ X_{LL}, X_{HL}, X_{LH} \} \) through a residual connection to maintain structural integrity. After processing, the modified \( X_{HH} \) subband and the original subbands  \( \{ X_{LL}, X_{HL}, X_{LH} \} \) are passed to the inverse wavelet transform (IWT), which reconstructs the features back into the spatial domain, restoring their original resolution and structural context. Finally, the output from the frequency branch is fused via element-wise addition with the spatial branch to form the complete refined feature map $F_j^{\mathrm{wsb}}$, which is then passed to the RGD blocks for image reconstruction.

\textbf{Prediction Layer:} The prediction layer consists of a \( 1 \times 1 \) Conv layer that maps the features to the final intensity image. This yields the reconstructed image \( \hat{I}_k \in \mathbb{R}^{H \times W \times 1} \). We do not use an activation function after this layer.

\subsection{Loss Functions} 
To train our model, we employ a combination of two loss functions to ensure good visual quality and temporally coherent results. Firstly, we use image reconstruction loss, i.e., learned perceptual image patch similarity (LPIPS) loss as employed in \cite{E2VID} to maintain natural statistics and visual quality in the reconstructed images. The loss function LPIPS measures the difference between the reconstructed image $\hat{I}_k$ and the corresponding GT image $I_k$ by comparing deep feature representations extracted using a VGG network \cite{vgg} pre-trained in ImageNet \cite{imagenet}. Formally, the perceptual reconstruction loss is defined as $\mathcal{L}^R_k = \ell(\hat{I}_k, I_k)$, where $\ell$ denotes the LPIPS distance. Secondly, we incorporate a temporal consistency loss, $\mathcal{L}_k^{TC}$, proposed in~\cite{E2VID}, to enforce smooth transitions between consecutive images in the reconstructed image sequence. This loss function utilizes GT optical flow maps, $F_{k-1}^k$, to warp the previous reconstructed image, $\hat{I}_{k-1}$, to the current image, $\hat{I}_k$, measuring the warping error. The temporal consistency loss is given by equation (\ref{eq:temporal_consistency}):
\begin{equation}
\mathcal{L}_k^{TC} = M_{k-1}^k \left\| \hat{I}_k - \mathcal{W}_{k-1}^k(\hat{I}_{k-1}) \right\|_1
\label{eq:temporal_consistency}
\end{equation}
\noindent where $\mathcal{W}_{k-1}^k(\hat{I}_{k-1})$ is the result of warping the reconstructed image at time step $k{-}1$ to time step $k$, using the optical flow, $F_{k-1}^k$, and $M_{k-1}^k = \exp\left(-\alpha \left\| I_k - \mathcal{W}_{k-1}^k(I_{k-1}) \right\|_2^2 \right)$ is a weighting term that accounts for occlusions. In our experiments, we set $\alpha = 50$ as done in \cite{E2VID,E2VID+,FireNet}. The final loss function, $\mathcal{L}$, is computed using equation (\ref{eq:final_loss}) as a weighted sum of the reconstruction loss and the temporal consistency loss across all time steps.
\begin{equation}
\mathcal{L} = \sum_{k=1}^{L} \mathcal{L}_k^R + \lambda_{TC} \sum_{k=L_0}^{L} \mathcal{L}_k^{TC}
\label{eq:final_loss}
\end{equation}
In our experiments, we set $L = 40$, which denotes the sequence length of consecutive images considered by our model during training. This length controls how many time steps the model unrolls to capture temporal dependencies. The temporal consistency loss weight, $\lambda_{TC}$, is empirically set to 5 to balance its influence on the reconstruction loss. Additionally, we set $L_0 = 2$, following the strategy used  in~\cite{E2VID}.
\section{Experiments and Results}
\textbf{Training Dataset:} We follow a common approach to generate a synthetic training dataset \cite{E2VID+}, using the ”Multiple-Object-2D” rendering engine of ESIM \cite{esim}. This engine simulates multiple foreground objects moving across a background image, employing various 2D motion properties. Background images were selected from the MSCOCO dataset\cite{mscoco}, while foreground objects were sourced from\cite{E2VID+}. The dataset consists of 280 sequences, each 10 seconds in length. The contrast threshold values for event generation ranged from 0.1 to 1.5. Each sequence in our training dataset comprises an event stream together with the corresponding GT images and optical flow maps produced at an average rate of 51Hz. The resolutions of event and image cameras are both $240 \times 180$ pixels to match the DAVIS240C sensor\cite{Davis240C}.

\textbf{Training Settings:} We implement the proposed model in PyTorch for training. The network is trained on 200 epochs with a batch size of 2 on an RTX 3080 GPU with 16GB of RAM. We utilize the Adam optimizer with an initial learning rate of $10^{-4}$, which decays by 10\% every 50 epochs. We also perform data augmentation with random crop and flip operations. The size of the crop was set to $128 \times 128$, and the probability of both vertical and horizontal flips was set to 0.5. To further enhance performance and mitigate overfitting, we incorporate noise augmentation, pause augmentation, and hot-pixel augmentation following the methods in\cite{E2VID+}.

\textbf{Testing Datasets:} To thoroughly assess our proposed model, we utilize sequences from five real-world datasets, each chosen for its distinct characteristics and significance in capturing various aspects of event-based video reconstruction:

\begin{enumerate}
    \item The event camera dataset (ECD)\cite{ecd} is recorded with a DAVIS240C sensor at a resolution of $240 \times 180$ pixels. It comprises seven indoor sequences that are pivotal for evaluating reconstructions in environments with 6-DOF camera motion and varying speeds.
    \item The high-quality frames (HQF)\cite{E2VID+} dataset is also recorded using two DAVIS240C sensors at the resolution of $240 \times 180$ pixels. It provides a variety of indoor and outdoor sequences with well-exposed and minimally blurred frames, crucial for benchmarking reconstruction quality in more controlled environments.
    \item The multi vehicle stereo event camera (MVSEC)\cite{mvsec} dataset is captured using a DAVIS346B sensor at a resolution of $346 \times 260$ pixels. It includes sequences recorded in real-world driving scenarios, featuring diverse lighting conditions and dynamic motion, making it valuable for evaluating reconstruction performance in complex and fast-changing outdoor scenes.
    \item To demonstrate the generalizability of our model on a higher-resolution sensor, we also utilize a high-speed and HDR dataset (recorded with Samsung DVS Gen3 with a resolution of $640 \times 480$ pixels)\cite{E2VID}. It includes challenging scenarios with strong light contrasts.
    \item The color event camera dataset (CED)\cite{ced}, captured with a ColorDAVIS346 sensor, provides color frames and events, allowing us to demonstrate our method’s ability to reconstruct color in scenes with vibrant textures and challenging lighting conditions.
\end{enumerate}
\input{table.tex}
\input{figure_all}

\input{figure_slow}
\textbf{Evaluation Metrics:} To quantitatively assess the proposed model, the following commonly used full reference metrics were used: peak signal-to-noise ratio (PSNR)  $[\uparrow]$, structural similarity index measure (SSIM)  $[\uparrow]$, and LPIPS  $[\downarrow]$.  Additionally, we utilized a no-reference image quality assessment metric, blind/referenceless image spatial quality evaluator (BRISQUE) $[\downarrow]$. This metric is used in the literature\cite{HyperE2VID} when the GT images are of low quality or are unavailable. 

\textbf{Comparison with SOTA Methods:} We compare our model with several SOTA methods for which publicly available code exists: the CNN-based models, E2VID\cite{E2VID}, FireNet\cite{FireNet}, SPADE-E2VID\cite{SPADE-E2VID}, FireNet+\cite{E2VID+}, E2VID+\cite{E2VID+}, HyperE2VID\cite{HyperE2VID}, CISTA-LSTC\cite{cista-lstc}, and the transformer-based method ET-Net\cite{ET-Net}. For fair evaluation, we re-trained all models on our synthetic dataset using the same settings as MSFET-E2V, except for E2VID and FireNet, which use pre-trained weights due to their original training on a different dataset. As the GT images of ECD are darker, we follow a common approach\cite{E2VID,cista-lstc} of normalizing the images into the range of $[0, 1]$ to enable comparison with reconstructed images without applying post-processing to the reconstructed images. In our work, we utilized the EVREAL framework\cite{evreal} to assess the reconstruction quality.
\input{figure_low_light}

Table \ref{Table:quant_results} and Fig. \ref{fig:visual_all} present the quantitative and visual evaluation of our proposed model compared to SOTA methods across three benchmark datasets: HQF, MVSEC, and ECD. For the HQF and MVSEC datasets, our model achieves a significant improvement across all evaluation metrics when compared to both CNN and the transformer-based approaches. This performance gains highlight the effectiveness of the proposed model that operates in both spatio-temporal and frequency domains. As shown in Fig. \ref{fig:visual_all} (rows 1–2, HQF dataset), most baseline methods fail to recover fine object details and exhibit prominent bleeding and ghosting artifacts. In contrast, our model produces more structurally consistent reconstructions with reduced artifacts, leading to higher SSIM and lower LPIPS values. Notably, in challenging indoor environments (e.g., row 3), where methods like CISTA-LSTC, E2VID+ and HyperE2VID exhibit severe distortions (e.g., unrealistic contrast and artifacts on the floor and drum surface), our method reconstructs intensity images with minimal blur and intensity closer to the GT. In the case of the ECD dataset, our model (16 Mparam) surpasses the existing transformer-based architecture (22 Mparam) in both LPIPS and SSIM and outperforms all CNN-based baselines across most metrics. While PSNR is slightly better for HyperE2VID, our model still offers comparable perceptual similarity. As evidenced in Fig. \ref{fig:visual_all} (rows 5, 6, and 7), previous methods show limited capability in capturing details, with more pronounced blur and ghosting. In contrast, our reconstruction maintains sharper contours and better texture preservation, contributing to improved perceptual and structural scores (see shapes and dynamic).
\input{figures_hdr}

We also evaluate the reconstruction capabilities of the proposed model in slow motion conditions, comparing it with the two best-performing baselines: ET-Net and CISTA-LSTC (see Table \ref{Table:quant_results}). In Fig. \ref{fig:slow_low}, we present results from the Desk\_Slow sequence of the HQF dataset, where the camera motion slows down and event generation diminishes, causing intensity information to vanish. Despite this, our model demonstrates strong robustness, preserving structural details with minimal artifacts, resulting in lower LPIPS and higher SSIM scores. While CISTA-LSTC also preserves structural content reasonably well, it introduces unrealistic artifacts and exaggerated contrast (particularly around text and object edges), which compromises visual fidelity and results in higher LPIPS scores.

To evaluate the performance of the proposed model under extremely low-light conditions, we test it on the MVSEC\_night subset of the MVSEC dataset, following the setup introduced by\cite{HyperE2VID}. As shown in Table~\ref{Table:quan_low}, we assess reconstruction quality using the BRISQUE metric, with corresponding visual results illustrated in Fig.~\ref{fig:low_light} (compared to the two best-performing baselines: HyperE2VID and ET-Net). Our model consistently achieves lower BRISQUE scores and better preserves scene details, indicating that the reconstructions are perceptually closer to real-world images.
\input{table_computational}

To validate the superiority of the proposed method on higher resolution and high speed motion scenarios, we also tested our model on high-speed and HDR dataset $(640 \times 480$ vs $240 \times 180)$. The GT images for this dataset are not available; therefore, we used BRISQUE evaluation metrics to evaluate the perceptual quality of reconstructed images. Table \ref{Table:quan_low} shows the quantitative results of the proposed method against SOTA methods. Our model achieves the lowest BRISQUE score, confirming that the reconstruction is closer to the natural images. This can be seen in Fig. \ref{fig:hdr}, where our model reconstructs high-quality and natural-looking images with fewer artifacts (see Sun) compared to the SOTA methods.
\input{figures_color}

Additionally, a comparison of computational resource consumption is presented in Table \ref{Table:params_speed}, where we report the average inference time in milliseconds (ms), per sequence for each method, evaluated across test datasets with varying resolutions. Model complexity is indicated by the number of parameters (in millions), and inference times are measured on an RTX 3080 GPU. Compared to the transformer-based ET-Net, our model achieves significant reductions (between 53\% and 73\%) in resource usage (highlighted in green), while maintaining competitive or superior reconstruction quality. This underscores the advantage of our frequency-enhanced self-attention mechanism, which enables both effective E2V reconstruction and efficient utilization of computational resources.

\input{figure_robust}
In Fig.~\ref{fig:color}, we also validate the color reconstruction capability of our model by comparing it against E2VID+, ET-Net, and HyperE2VID. Results from CISTA-LSTC are omitted, as the method fails to produce satisfactory outcomes on this dataset. For color reconstruction, we adopt the procedure introduced by \cite{E2VID}. The results clearly demonstrate that our model not only assigns natural-looking colors to images without introducing artifacts but also successfully reconstructs fine details under challenging conditions, such as scenes with high dynamic range (e.g., Indoor Dark reconstruction).

To further evaluate the robustness of the proposed model, we follow the evaluation protocol outlined in \cite{HyperE2VID} where we assess the effect of using different event grouping strategies. First, we investigate the fixed temporal window grouping scenario by conducting ten experiments with varying integration windows, ranging from 10 ms to 100 ms. This allows us to analyze how the temporal resolution impacts the reconstruction quality. Second, we evaluate the performance of the proposed model under a fixed number of events grouping, conducting another nine experiments with event counts ranging from 5K to 45K. For both scenarios, we compute the mean LPIPS scores on the ECD dataset and compare our model against three best-performing methods: ET-Net, CISTA-LSTC, and HyperE2VID. As shown in Fig. \ref{fig:robust}, the proposed method consistently achieves superior reconstruction quality across a wide range of event grouping conditions. The only exceptions are the 80 ms and 30k event settings, where CISTA-LSTC obtains slightly better LPIPS scores.
\input{tables_ablation}

\section{Ablation Study}
\label{sec:ablation}
\subsection{Investigation on Network Modules} 
We design three additional model variants to analyze the role of each component in our proposed model.

\textbf{Model I:} In this baseline variant, the DownConv blocks remain unchanged. However, the frequency domain pathway is removed from the CDAMs, and all attention components $(Q$, $K$, and $V)$ are derived solely from the spatio-temporal domain. This variant also omits the proposed WSB, instead using standard residual skip connections between DownConv and RGD blocks. In RGD blocks, the reconstruction is performed without RB. As shown in Table~\ref{Table:ablation_1} (Model I), this leads to a significant drop in SSIM and LPIPS on all event datasets, highlighting the impact of excluding frequency information and advanced skip connections. Visual results in Fig.~\ref{fig:modules} further emphasize the resulting loss of detail and over-smoothed textures. Although Model I maintains structural consistency, it lacks textural sharpness and fine-detail fidelity.
\input{table_ablation_II}

\textbf{Model II:} This variant reintroduces the frequency domain pathway in the CDAM module, enabling the model to exploit both spatio-temporal and frequency domain cues. WSB and RB remain excluded. The performance improvement over Model II confirms the effectiveness of frequency-based attention in capturing structural and textural features that are essential for high-fidelity reconstructions.

\textbf{Model III:} Building on Model II, we incorporate RBs into RGD blocks to enhance representation learning during reconstruction. These blocks help preserve feature flow and facilitate deeper learning, yielding further improvements in quantitative scores. WSB is still not included, allowing for an isolated evaluation of RBs contribution.

Finally, by introducing the WSB into the architecture, we arrive at the complete proposed MSFET-E2V model.  As evidenced in Fig. \ref{fig:modules}, this full configuration yields the best performance across all metrics and produces visually sharper, more detailed reconstructions
\subsection{Investigation on Frequency Domain}
In this experiment, we design models with different choices of subbands $(X_{LL}, X_{HL}, X_{LH}, X_{HH})$ to see their role in CDAM. Here, we introduce two variants of CDAM.

CDAM\_LL processes only the LF subband, $X_{LL}$, in the frequency domain path, omitting all HF components. In this variant, the input undergoes DWT, but only the $LL$ subband is passed through RB, while the HF components are discarded. The rest of the CDAM module, including the spatio-temporal path, remains unchanged. In contrast, CDAM\_HF processes only the HF subbands $(X_{HL}, X_{LH}, X_{HH})$) in the frequency domain path of CDAM. As shown in Table~\ref{Table:ablation_cdam}, the removal of HF subbands leads to a notable degradation in performance across all datasets. Despite the spatio-temporal branch retaining structural cues, the absence of HF detail leads to a noticeable drop in SSIM and PSNR, with LPIPS scores indicating perceptual degradation in CDAM\_LL model. As shown in Fig.~\ref{fig:ablation_cdam}, the outputs are overly smoothed and lack sharp textures, highlighting the limited ability of $X_{LL}$ alone to capture fine-grained variations.

In contrast, CDAM\_HF  better preserves local textures and edges, resulting in improved perceptual quality and sharper reconstructions. However, it lacks the global context and structural coherence provided by $X_{LL}$, which is reflected in relatively lower SSIM compared to the full model (MSFET-E2V). These observations confirm that both LF and HF components play complementary roles in enhancing reconstruction quality in the proposed CDAM block.

\subsection{Investigation on Encoder-Decoder Depth}
In this experiment we analyze the impact of varying the encoder-decoder depth $(d)$ of proposed model on E2V reconstruction performance. As shown in Table~\ref{Table:depth_ablation}, increasing depth from $d=1$ to $d=3$ consistently improves all performance metrics (PSNR$\uparrow$, SSIM $\uparrow$, and LPIPS $\downarrow$) across ECD, HQF, and MVSEC datasets, highlighting the role of deeper representations in learning spatial-frequency correlations. While depth $d=4$ offers marginal gains in HQF, it underperforms slightly on ECD and MVSEC. To meet the demand for efficient models that use transformer-based self-attention for global feature extraction, we focus on minimizing memory usage. Since increasing the depth beyond $d=3$ offers no notable gains, we adopt $d=3$ as the optimal configuration for our final model.
\subsection{Investigation on the Choice of Temporal Bins}
We investigate the influence of the number of voxel bins used to encode the event stream on the quality of E2V reconstruction. As shown in Table \ref{Table:voxel_bins}, we evaluate our model across three benchmark datasets. The results show that increasing the number of voxel bins from 2 to 5 progressively improves reconstruction performance. This is because a higher number of bins allows more precise temporal discretizations, thereby enabling the model to extract richer motion and structure cues from asynchronous events. However, beyond 5 bins (e.g., 6, 8, or 10), we found that the values produced no significant differences. Particularly, the performance of proposed model decreases when bin number was set to 10. This could be because the higher number of bins provides more temporal resolution, potentially capturing finer details of motion, but it can also lead to sparse event representations, where many bins contain little or no information. Given these trade-offs and in alignment with most of the prior works in the literature\cite{E2VID,E2VID+, HyperE2VID, ET-Net}, we select 5 as the optimal number of bins. This configuration balances temporal resolution and computational efficiency, leading to the best reconstruction quality among the tested settings.

\section{Conclusion}
In this paper, we propose MSFET-E2V, a novel multiscale frequency-enhanced transformer model for high-quality event-to-video (E2V) reconstruction. Unlike existing methods, MSFET-E2V operates across three complementary domains, i.e., spatial, temporal, and frequency, to fully exploit the rich structure of event data. Central to our model is the proposed cross-domain attention module (CDAM), which effectively captures both local and global dependencies by utilizing convolutional recurrent networks in the spatio-temporal domain. To further enhance fine-grained detail preservation, we incorporate the discrete wavelet transform (DWT) within CDAM for frequency-aware feature representation. Additionally, we introduce the wavelet-enhanced skip block (WSB), which processes shallow encoder features in both spatial and frequency domains before fusing them with the features produced by reconstruction blocks. This design significantly reduces artifacts and improves image fidelity. We also address a major limitation of existing transformer-based E2V models, their high computational cost and restricted spatial-domain attention. Our MSFET-E2V model achieves superior reconstruction quality while reducing inference time by up to approximately 70\%, demonstrating that efficient, high-quality video reconstruction is feasible with reduced memory consumption.

\bibliographystyle{IEEEtran}  
\bibliography{references}  

\end{document}

%% file: table.tex
\begin{table*}[ht]
\centering
\renewcommand{\arraystretch}{1.2}
\setlength{\tabcolsep}{6pt}
\caption{Quantitative comparison on ECD, HQF, and MVSEC datasets. Best results are in \textbf{bold}, second-best are \underline{underlined}.}
\vspace{-2mm}
\begin{tabular}{llccc|ccc|ccc}
\toprule
\multirow{2}{*}{\textbf{Architecture Type}} & \multirow{2}{*}{\textbf{Methods}} & \multicolumn{3}{c|}{\textbf{ECD}} & \multicolumn{3}{c|}{\textbf{HQF}} & \multicolumn{3}{c}{\textbf{MVSEC}} \\
& & PSNR$\uparrow$ & SSIM$\uparrow$ & LPIPS$\downarrow$ & PSNR$\uparrow$ & SSIM$\uparrow$ & LPIPS$\downarrow$ & PSNR$\uparrow$ & SSIM$\uparrow$ & LPIPS$\downarrow$ \\
\midrule
\multirow{7}{*}{CNN-based}
& E2VID\cite{E2VID}        & 12.64 & 0.450 & 0.322 & 10.54 & 0.468 & 0.371 & 6.98  & 0.241 & 0.644 \\
& FireNet\cite{FireNet}      & 12.33 & 0.479 & 0.320 & 10.31 & 0.423 & 0.441 & 5.42  & 0.177 & 0.700 \\
& SPADE-E2VID\cite{SPADE-E2VID}   & 11.93 & 0.442 & 0.397 & 11.21 & 0.400 & 0.502 & 8.79  & 0.246 & 0.599 \\
& FireNet+\cite{E2VID+}      & 13.41 & 0.491 & 0.416 & 13.98 & 0.471 & 0.314 & 6.93  & 0.224 & 0.598 \\
& E2VID+\cite{E2VID+}        & 13.44 & 0.482 & 0.317 & 12.83 & 0.414 & 0.421 & 9.17  & 0.228 & 0.618 \\
& HyperE2VID\cite{HyperE2VID}   & \textbf{14.10}& 0.494 & 0.326 & 12.61 & 0.460 & 0.379 & 9.61  & 0.278 & \underline{0.558} \\
& CISTA-LSTC\cite{cista-lstc}   & 13.04 & \underline{0.526} & 0.362 & 12.40 & \underline{0.511} & 0.343 & 6.32  & 0.209 & 0.618 \\
\midrule
\multirow{2}{*}{Transformer-based}
& ET-Net\cite{ET-Net}        & 13.52& 0.521 & \underline{0.275} & \underline{14.28} & 0.502 & \underline{0.331} & \underline{10.15} & \underline{0.311} & 0.568 \\
& MSFET-E2V              & \underline{13.98}& \textbf{0.584} & \textbf{0.252} & \textbf{14.82} & \textbf{0.548} & \textbf{0.291} & \textbf{11.42} & \textbf{0.362} & \textbf{0.501} \\
\bottomrule
\end{tabular}
\label{Table:quant_results}
\end{table*}
\begin{table}[t]
\centering
\caption{BRISQUE (lower is better) score comparison on MVSEC\_night and high-speed HDR datasets. Best results are in \textbf{bold}, second-best are \underline{underlined}.}
\vspace{-2mm}
\label{Table:quan_low}
\begin{tabular}{@{}lcc@{}}
\toprule
\textbf{Methods} & \textbf{MVSEC\_night} & \textbf{high-speed and HDR} \\
\midrule
E2VID\cite{E2VID}         & 19.16 & 22.83 \\
FireNet\cite{FireNet}       & 20.45 & 21.32 \\
SPADE-E2VID\cite{SPADE-E2VID}   & 24.01 & 27.83 \\
FireNet+\cite{E2VID+}     & 16.98 & 18.17 \\
E2VID+\cite{E2VID+}      & 18.30 & 13.42 \\
HyperE2VID\cite{HyperE2VID}    & 16.59 & \underline{11.89} \\
CISTA-LSTC\cite{cista-lstc}   & 32.53 & 38.74 \\
ET-Net\cite{ET-Net}        & \underline{16.17} & 18.47 \\
MSFET-E2V   & \textbf{9.32} & \textbf{7.40} \\
\bottomrule
\end{tabular}
\vspace{-4mm}
\end{table}

%% file: figure_all.tex
\begin{figure*}[t]
    \renewcommand{\arraystretch}{0.8}
    \setlength{\tabcolsep}{1pt}
    \begin{adjustbox}{max width=\textwidth,center}
    \begin{tabular}{>{\centering\arraybackslash}m{0.25cm} *{10}{>{\centering\arraybackslash}m{0.10\textwidth}}}
        & \scriptsize Event Frame & \scriptsize SPADE-E2VID & \scriptsize FireNet & \scriptsize FireNet+ & \scriptsize E2VID+ & \scriptsize HyperE2VID & \scriptsize CISTA-LSTC & \scriptsize ET-Net & \scriptsize MSFET-E2V & \scriptsize GT \\
        \rotatebox[origin=c]{90}{\parbox{0.6cm}{\centering \scriptsize Bike\_Bay}} &
        \includegraphics[width=\linewidth]{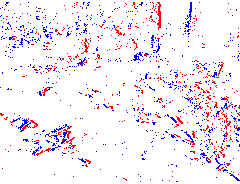} &
        \includegraphics[width=\linewidth]{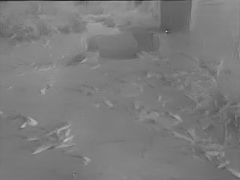} &
        \includegraphics[width=\linewidth]{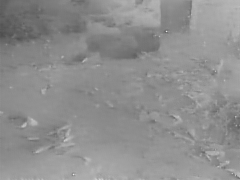} &
        \includegraphics[width=\linewidth]{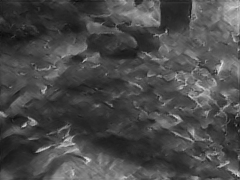} &
        \includegraphics[width=\linewidth]{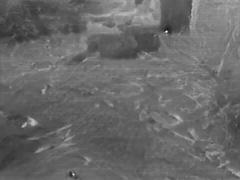} &
        \includegraphics[width=\linewidth]{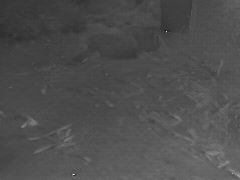} &
        \includegraphics[width=\linewidth]{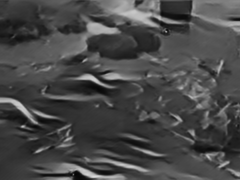} &
        \includegraphics[width=\linewidth]{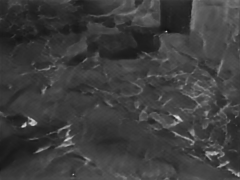} &
        \includegraphics[width=\linewidth]{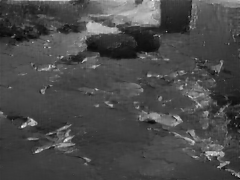} &
        \includegraphics[width=\linewidth]{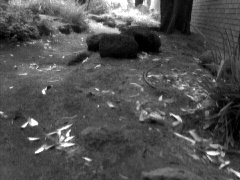} \\        
        \rotatebox[origin=c]{90}{\parbox{0.6cm}{\centering \scriptsize Desk\_Slow}} &
        \includegraphics[width=\linewidth]{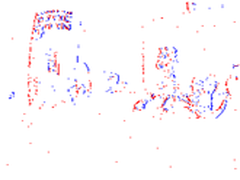} &
        \includegraphics[width=\linewidth]{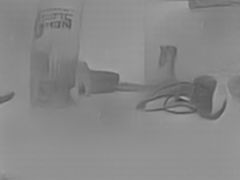} &
        \includegraphics[width=\linewidth]{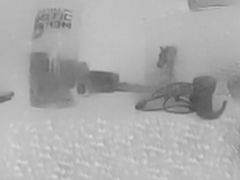} &
        \includegraphics[width=\linewidth]{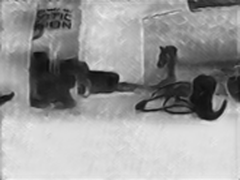} &
        \includegraphics[width=\linewidth]{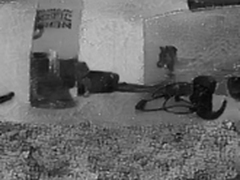} &
        \includegraphics[width=\linewidth]{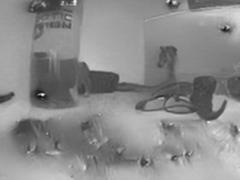} &
        \includegraphics[width=\linewidth]{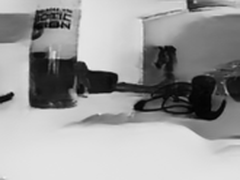} &
        \includegraphics[width=\linewidth]{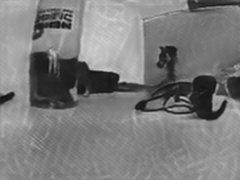} &
        \includegraphics[width=\linewidth]{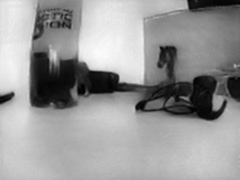} &
        \includegraphics[width=\linewidth]{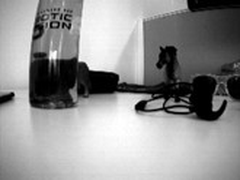} \\
        \rotatebox[origin=c]{90}{\parbox{0.6cm}{\centering \scriptsize Indoor}} &
        \includegraphics[width=\linewidth]{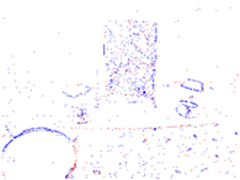} &
        \includegraphics[width=\linewidth]{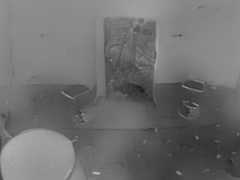} &
        \includegraphics[width=\linewidth]{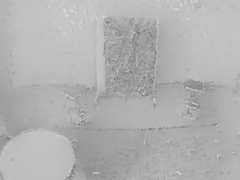} &
        \includegraphics[width=\linewidth]{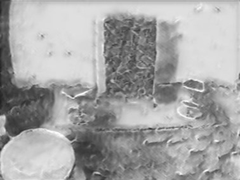} &
        \includegraphics[width=\linewidth]{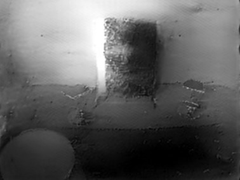} &
        \includegraphics[width=\linewidth]{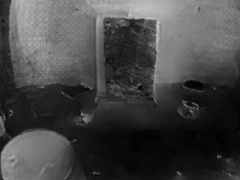} &
        \includegraphics[width=\linewidth]{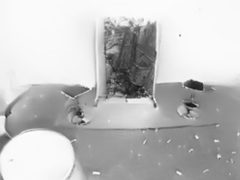} &
        \includegraphics[width=\linewidth]{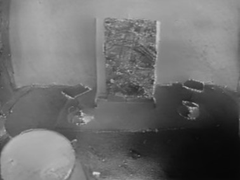} &
        \includegraphics[width=\linewidth]{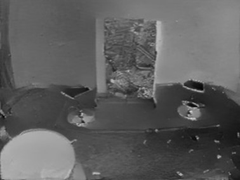} &
        \includegraphics[width=\linewidth]{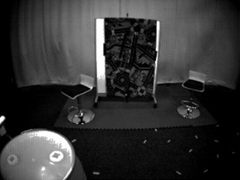} \\
        \rotatebox[origin=c]{90}{\parbox{0.6cm}{\centering \scriptsize Outdoor}} &
        \includegraphics[width=\linewidth]{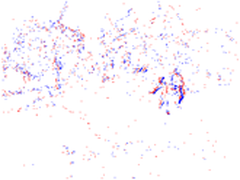} &
        \includegraphics[width=\linewidth]{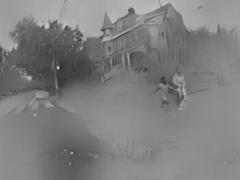} &
        \includegraphics[width=\linewidth]{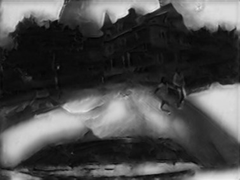} &
        \includegraphics[width=\linewidth]{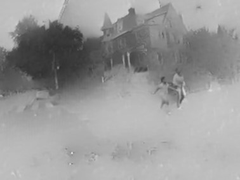} &
        \includegraphics[width=\linewidth]{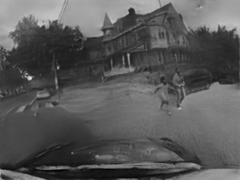} &
        \includegraphics[width=\linewidth]{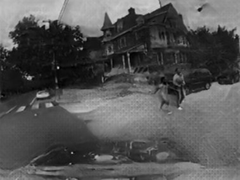} &
        \includegraphics[width=\linewidth]{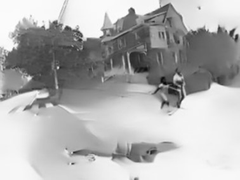} &
        \includegraphics[width=\linewidth]{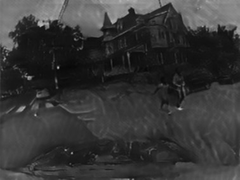} &
        \includegraphics[width=\linewidth]{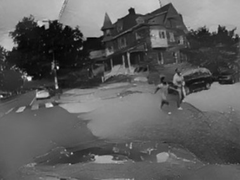} &
        \includegraphics[width=\linewidth]{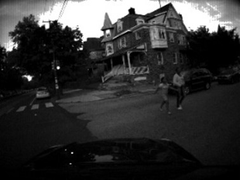} \\
        \rotatebox[origin=c]{90}{\parbox{0.6cm}{\centering \scriptsize Dynamic}} &
        \includegraphics[width=\linewidth]{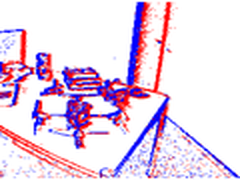} &
        \includegraphics[width=\linewidth]{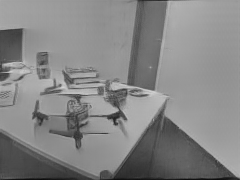} &
        \includegraphics[width=\linewidth]{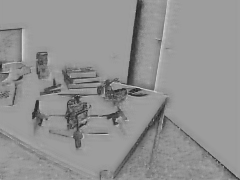} &
        \includegraphics[width=\linewidth]{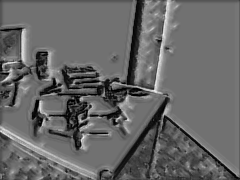} &
        \includegraphics[width=\linewidth]{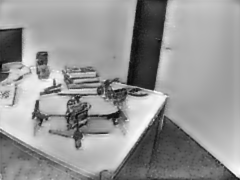} &
        \includegraphics[width=\linewidth]{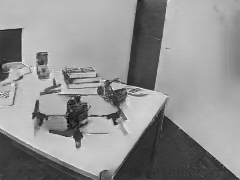} &
        \includegraphics[width=\linewidth]{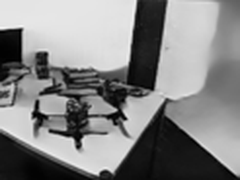} &
        \includegraphics[width=\linewidth]{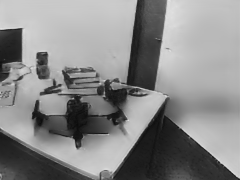} &
        \includegraphics[width=\linewidth]{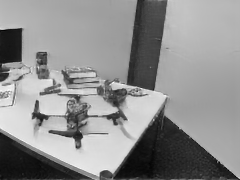} &
        \includegraphics[width=\linewidth]{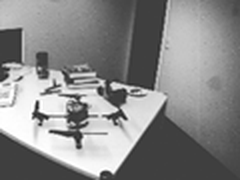} \\
        \rotatebox[origin=c]{90}{\parbox{0.6cm}{\centering \scriptsize Poster}} &
        \includegraphics[width=\linewidth]{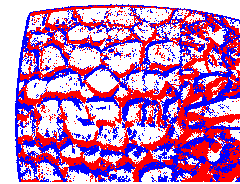} &
        \includegraphics[width=\linewidth]{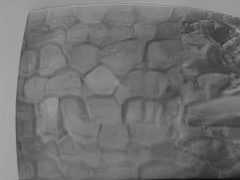} &
        \includegraphics[width=\linewidth]{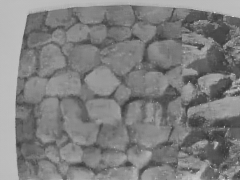} &
        \includegraphics[width=\linewidth]{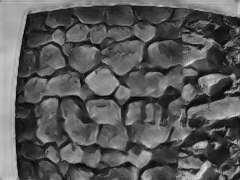} &
        \includegraphics[width=\linewidth]{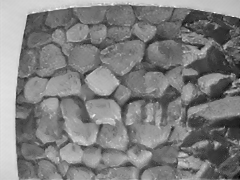} &
        \includegraphics[width=\linewidth]{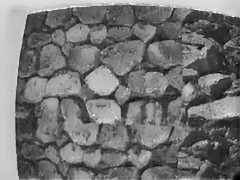} &
        \includegraphics[width=\linewidth]{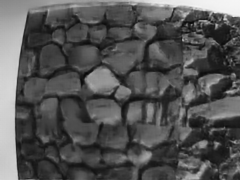} &
        \includegraphics[width=\linewidth]{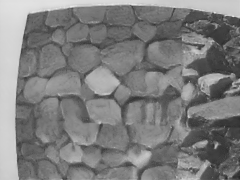} &
        \includegraphics[width=\linewidth]{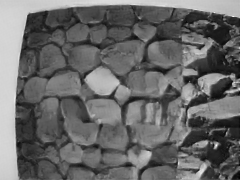} &
        \includegraphics[width=\linewidth]{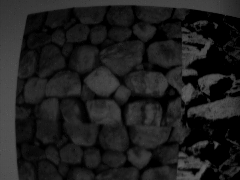} \\
        \rotatebox[origin=c]{90}{\parbox{0.6cm}{\centering \scriptsize Shapes}} &
        \includegraphics[width=\linewidth]{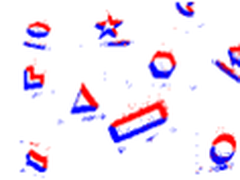} &
        \includegraphics[width=\linewidth]{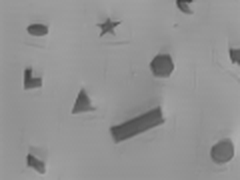} &
        \includegraphics[width=\linewidth]{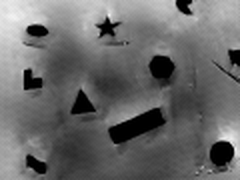} &
        \includegraphics[width=\linewidth]{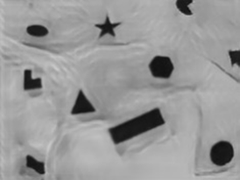} &
        \includegraphics[width=\linewidth]{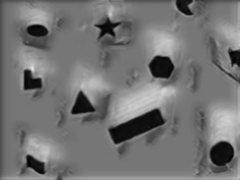} &
        \includegraphics[width=\linewidth]{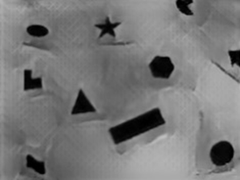} &
        \includegraphics[width=\linewidth]{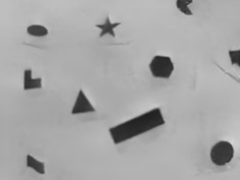} &
        \includegraphics[width=\linewidth]{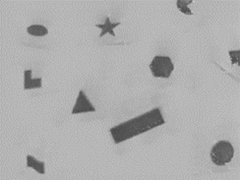} &
        \includegraphics[width=\linewidth]{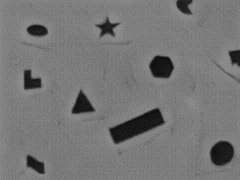} &
        \includegraphics[width=\linewidth]{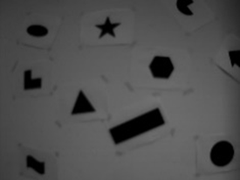} \\
    \end{tabular}
    \end{adjustbox}
    \caption{Visual results on HQF (rows 1,2), MVSEC (rows 3,4) and ECD (last three rows) datasets. Overall, our model provides high-quality reconstruction with minimal ghosting, bleeding and blur artifacts.}
    \label{fig:visual_all}
\vspace{-2mm}
\end{figure*}

%% file: figure_slow.tex
\begin{figure*}[ht]
\centering
\setlength{\tabcolsep}{1pt} 
\renewcommand{\arraystretch}{0.6} 
\begin{tabular}{@{} >{\centering\arraybackslash}m{1.9cm} *{6}{>{\centering\arraybackslash}m{0.13\textwidth}} @{}}
& \small Slow & \small Very Slow & \small Static & \small Static & \small Very Slow & \small Very Slow \\
\rotatebox[origin=c]{90}{\parbox[c][3.3cm][c]{1.9cm}{\centering   \mbox{CISTA-LSTC}}} &
\includegraphics[valign=m,width=\linewidth]{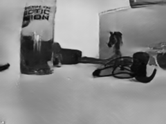} &
\includegraphics[valign=m,width=\linewidth]{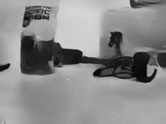} &
\includegraphics[valign=m,width=\linewidth]{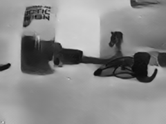} &
\includegraphics[valign=m,width=\linewidth]{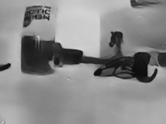} &
\includegraphics[valign=m,width=\linewidth]{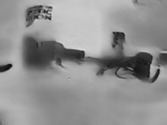} &
\includegraphics[valign=m,width=\linewidth]{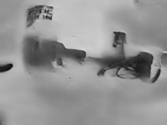} \\[-1pt]
&   0.24/0.66 &   0.29/0.64 &   0.30/0.63 &   0.30/0.63 &   0.45/0.57 &   0.45/0.56 \\[-1pt]
\rotatebox[origin=c]{90}{\parbox[c][3.3cm][c]{1.9cm}{\centering   \mbox{ET-Net}}} &
\includegraphics[valign=m,width=\linewidth]{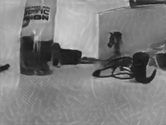} &
\includegraphics[valign=m,width=\linewidth]{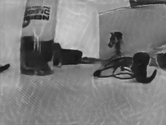} &
\includegraphics[valign=m,width=\linewidth]{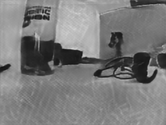} &
\includegraphics[valign=m,width=\linewidth]{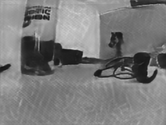} &
\includegraphics[valign=m,width=\linewidth]{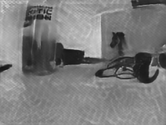} &
\includegraphics[valign=m,width=\linewidth]{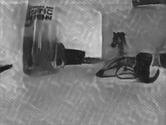} \\[-1pt]
&   0.31/0.49 &   0.31/0.49 &   0.31/0.49 &   0.31/0.48 &   0.38/0.46 &   0.38/0.45 \\[-1pt]
\rotatebox[origin=c]{90}{\parbox[c][3.3cm][c]{1.9cm}{\centering   \mbox{MSFET-E2V}}} &
\includegraphics[valign=m,width=\linewidth]{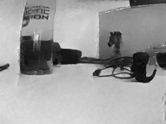} &
\includegraphics[valign=m,width=\linewidth]{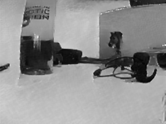} &
\includegraphics[valign=m,width=\linewidth]{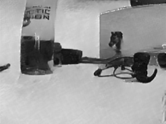} &
\includegraphics[valign=m,width=\linewidth]{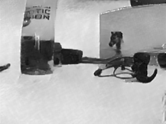} &
\includegraphics[valign=m,width=\linewidth]{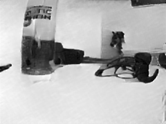} &
\includegraphics[valign=m,width=\linewidth]{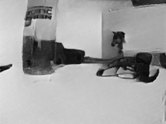} \\[-1pt]
&   \textbf{0.17/0.67} &   \textbf{0.19/0.67} &   \textbf{0.18/0.66} &   \textbf{0.18/0.65} &   \textbf{0.24/0.64} &   \textbf{0.22/0.65} \\
\end{tabular}
\caption{ Qualitative comparison in slow-motion conditions across different temporal speeds. Values below each image represent LPIPS/SSIM scores where \textbf{bold }values indicate the best-performing model.}
\label{fig:slow_low}
\vspace{-2mm}
\end{figure*}

%% file: figure_low_light.tex
\begin{figure*}[ht]
\centering
\setlength{\tabcolsep}{1pt} 
\renewcommand{\arraystretch}{0.6} 
\begin{tabular}{@{} >{\centering\arraybackslash}m{1.9cm} *{5}{>{\centering\arraybackslash}m{0.13\textwidth}} @{}}
\rotatebox[origin=c]{90}{\parbox[c][3.3cm][c]{1.9cm}{\centering    \mbox{HyperE2VID}}} &
\includegraphics[valign=m,width=\linewidth]{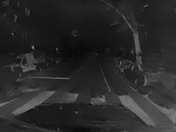} &
\includegraphics[valign=m,width=\linewidth]{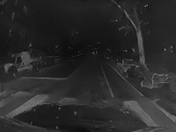} &
\includegraphics[valign=m,width=\linewidth]{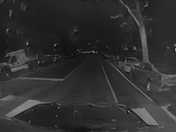} &
\includegraphics[valign=m,width=\linewidth]{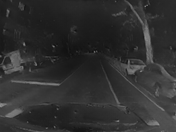} &
\includegraphics[valign=m,width=\linewidth]{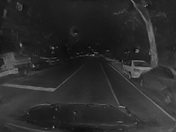} \\[-1pt]
&    21.78 &    21.64 &    20.54 &    20.65 &    19.87 \\[-1pt]
\rotatebox[origin=c]{90}{\parbox[c][3.3cm][c]{1.9cm}{\centering    \mbox{ET-Net}}} &
\includegraphics[valign=m,width=\linewidth]{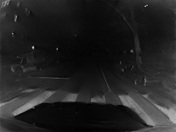} &
\includegraphics[valign=m,width=\linewidth]{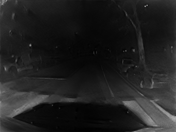} &
\includegraphics[valign=m,width=\linewidth]{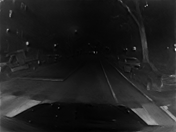} &
\includegraphics[valign=m,width=\linewidth]{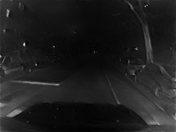} &
\includegraphics[valign=m,width=\linewidth]{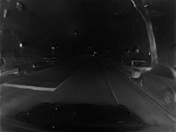} \\[-1pt]
&    22.87 &    22.42 &    23.71&    24.63 &    24.63 \\[-1pt]
\rotatebox[origin=c]{90}{\parbox[c][3.3cm][c]{1.9cm}{\centering    \mbox{MSFET-E2V}}} &
\includegraphics[valign=m,width=\linewidth]{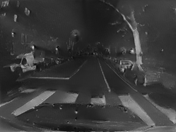} &
\includegraphics[valign=m,width=\linewidth]{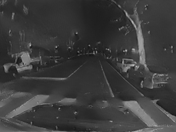} &
\includegraphics[valign=m,width=\linewidth]{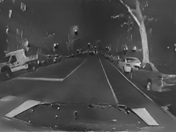} &
\includegraphics[valign=m,width=\linewidth]{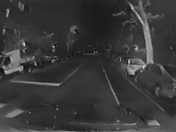} &
\includegraphics[valign=m,width=\linewidth]{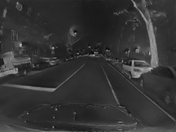} \\[-1pt]
&    \textbf{19.35} &    \textbf{19.64} &    \textbf{19.82} &    \textbf{19.84} &    \textbf{18.34} \\
\end{tabular}
\caption{Qualitative comparison in low-light scenario. Values below each show BRISQUE scores to support quantitative differences in reconstruction quality. Best values are shown in \textbf{bold}.}
\label{fig:low_light}
\vspace{-4mm}
\end{figure*}

%% file: figures_hdr.tex
\begin{figure}[ht]
\centering
\setlength{\tabcolsep}{0.5pt} 
\renewcommand{\arraystretch}{0.8} 
\begin{tabular}{@{} >{\centering\arraybackslash}m{0.3cm} *{3}{>{\centering\arraybackslash}m{0.13\textwidth}} @{}}
&  Sun &  Selfie &  Tunnel \\
\rotatebox[origin=c]{90}{\parbox[c][0.1cm][c]{1.9cm}{\centering  \mbox{HyperE2VID}}} &
\includegraphics[valign=m,width=\linewidth]{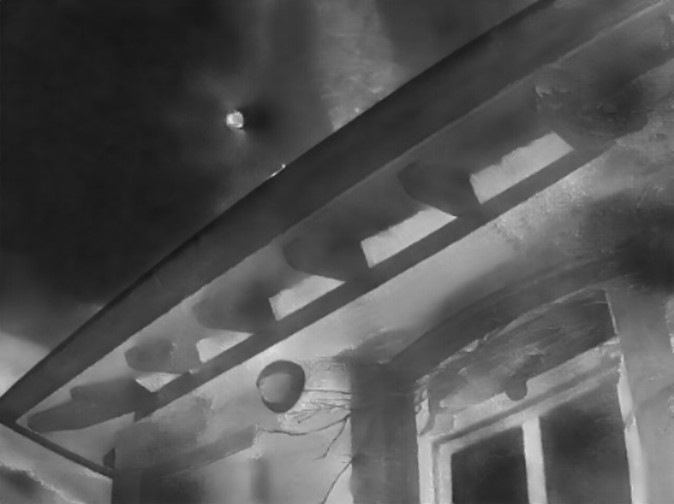} &
\includegraphics[valign=m,width=\linewidth]{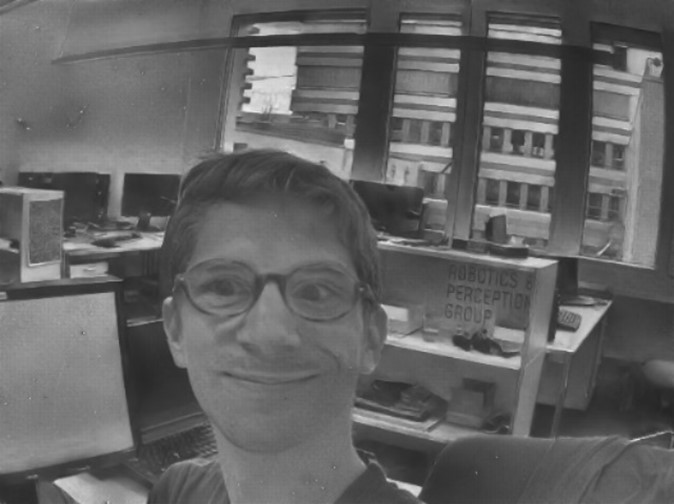} &
\includegraphics[valign=m,width=\linewidth]{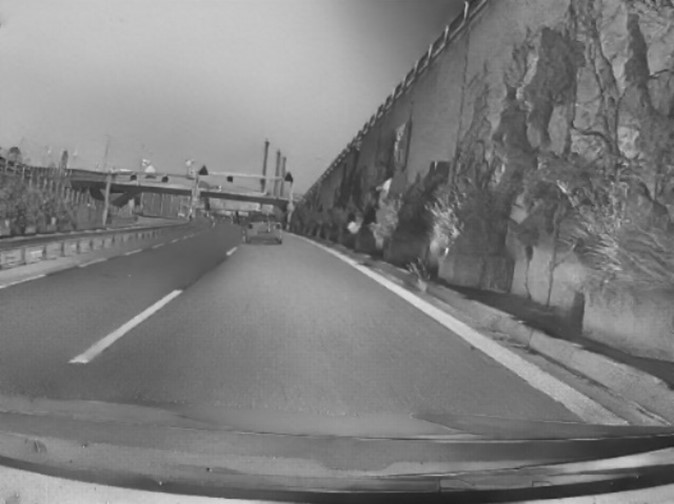} \\[-1pt]
&   18.10 &   32.55 &   6.90 \\[-1pt]
\rotatebox[origin=c]{90}{\parbox[c][0.1cm][c]{1.9cm}{\centering  \mbox{ET-Net}}} &
\includegraphics[valign=m,width=\linewidth]{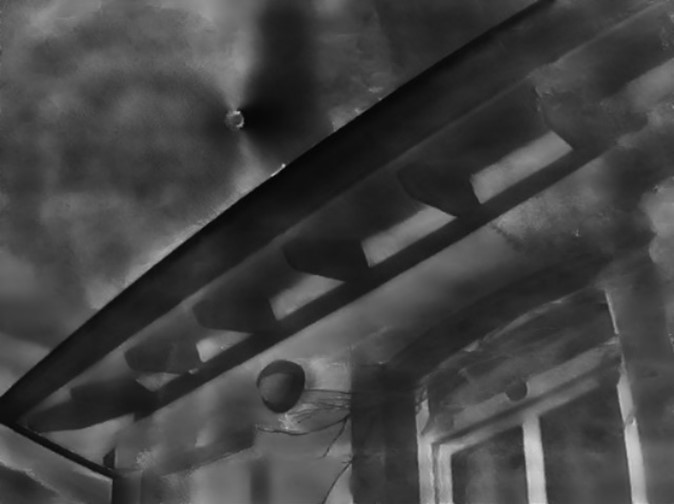} &
\includegraphics[valign=m,width=\linewidth]{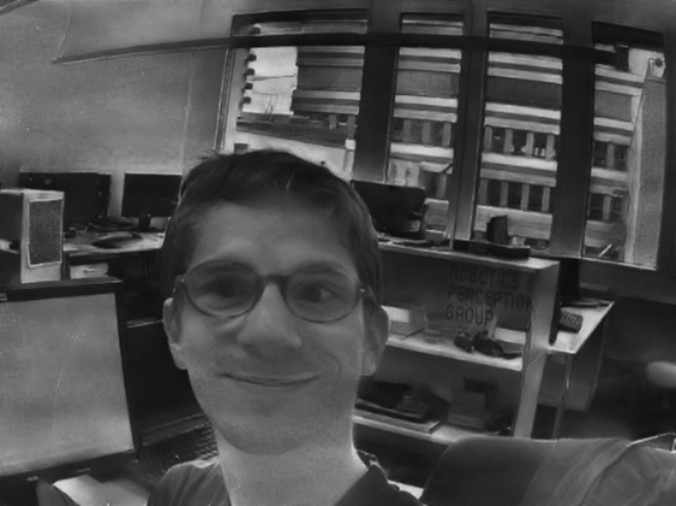} &
\includegraphics[valign=m,width=\linewidth]{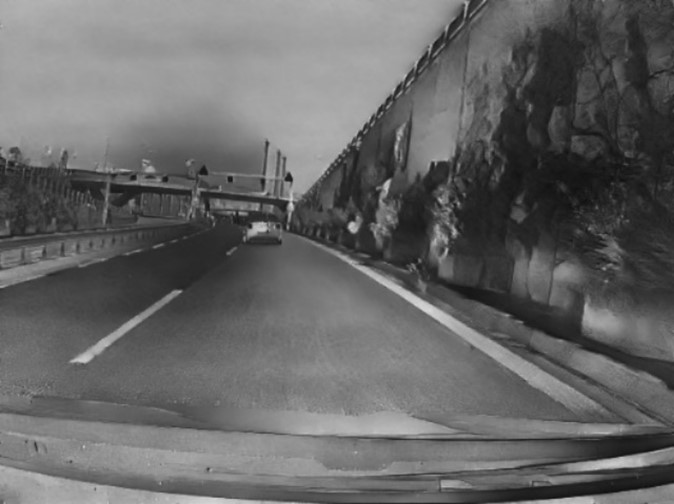} \\[-1pt]
&  43.86 &   18.30 &   9.45  \\[-1pt]
\rotatebox[origin=c]{90}{\parbox[c][0.1cm][c]{1.9cm}{\centering  \mbox{MSFET-E2V}}} &
\includegraphics[valign=m,width=\linewidth]{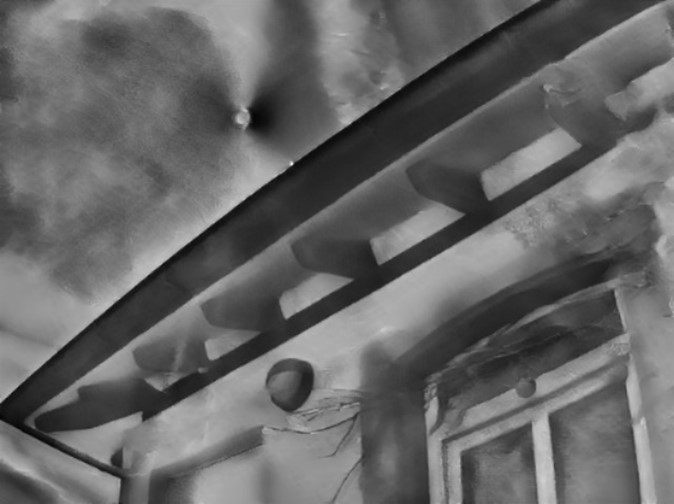} &
\includegraphics[valign=m,width=\linewidth]{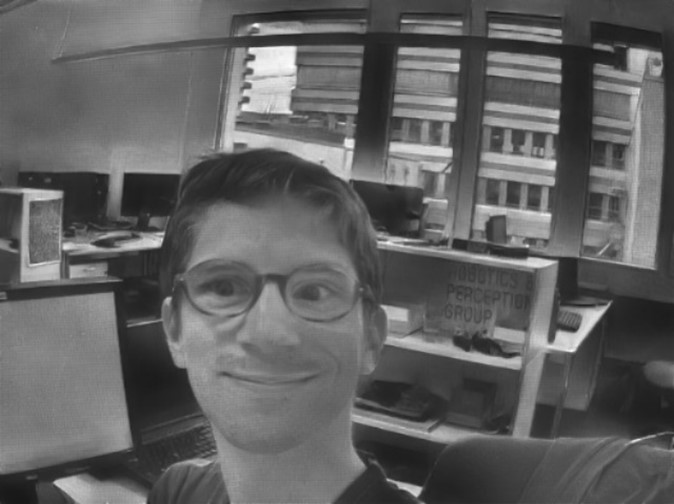} &
\includegraphics[valign=m,width=\linewidth]{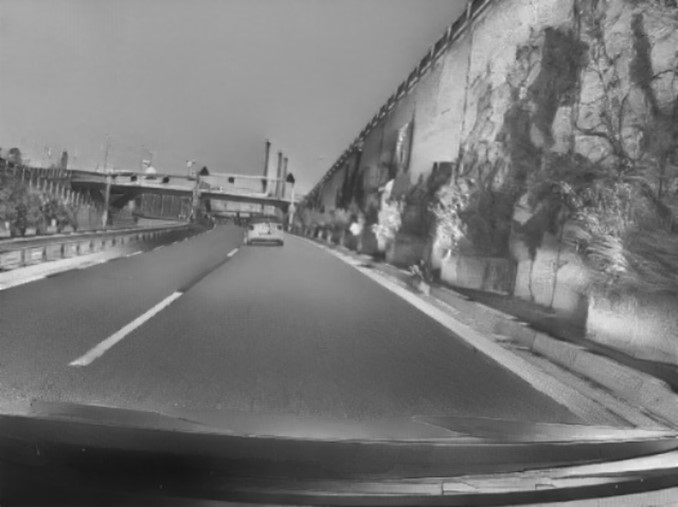} \\[-1pt]
&   \textbf{9.91} &   \textbf{5.12} &   \textbf{2.87} \\
\end{tabular}
\caption{Qualitative comparison under challenging conditions. Below each image are BRISQUE scores (lower is better). \textbf{Bold} values represent best performance.}
\label{fig:hdr}

\end{figure}

%% file: table_computational.tex
\begin{table}[t]
\centering
\caption{Comparison of model size and GPU inference speed across different input resolutions.}
\label{Table:params_speed}
\vspace{-2mm}
\begin{tabular}{@{}lccccc@{}}
\toprule
\multirow{2}{*}{\textbf{Methods}} & \textbf{Parameters} & \multicolumn{3}{c}{\textbf{GPU Inference Speed (ms)}} \\
\cmidrule(lr){3-5}
& (Millions) & 240×180 & 340×260 & 640×480 \\
\midrule
E2VID+\cite{E2VID+}       & 10.71 & 5.7   & 10.1 & 30.0 \\
FireNet+\cite{E2VID+}     & 0.04  & 0.6   & 3.7   & 10.1 \\
SPADE-E2VID\cite{SPADE-E2VID}  & 12.49 & 16.1  & 24.3  & 38.2 \\
HyperE2VID\cite{HyperE2VID}   & 10.15 & 6.6   & 12.9 & 22.0 \\
CISTA-LSTC\cite{cista-lstc}   & 10.40 & 2.0   & 24.3  & 52.9 \\
\rowcolor{lightgreen}
ET-Net\cite{ET-Net}       & 22.18 & 32.1  & 56.2  & 130.8 \\
\rowcolor{lightgreen}
MSFET-E2V           & 16.71 & 16.8  & 26.4  & 35.73 \\
\bottomrule
\end{tabular}
\vspace{-4mm}
\end{table}

%% file: figures_color.tex
\begin{figure}[ht]
\setlength{\tabcolsep}{0.5pt} 
\renewcommand{\arraystretch}{0.8} 
\resizebox{\columnwidth}{!}{%
\begin{tabular}{@{} >{\centering\arraybackslash}m{0.12\columnwidth} *{4}{>{\centering\arraybackslash}m{0.29\columnwidth}} @{}}
&  Simple Wire &  Jenga Block &  Indoor Dark & \ Indoor Kitchen \\[2pt]
\rotatebox[origin=c]{90}{\parbox[c][1.8cm][c]{1.0cm}{\centering  GT}} &
\includegraphics[width=\linewidth]{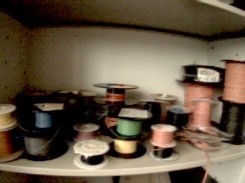} &
\includegraphics[width=\linewidth]{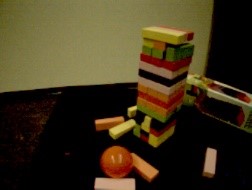} &
\includegraphics[width=\linewidth]{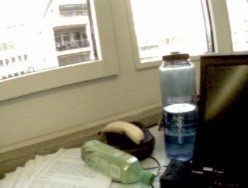} &
\includegraphics[width=\linewidth]{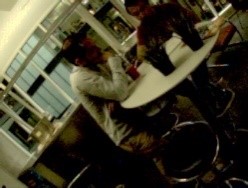} \\[2pt]
\rotatebox[origin=c]{90}{\parbox[c][1.8cm][c]{1.1cm}{\centering  \mbox{MSFET-E2V}}} &
\includegraphics[width=\linewidth]{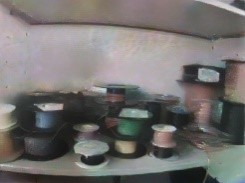} &
\includegraphics[width=\linewidth]{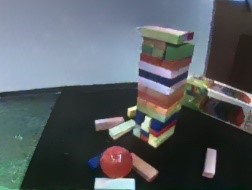} &
\includegraphics[width=\linewidth]{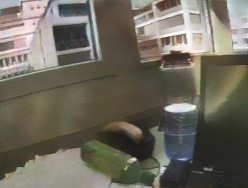} &
\includegraphics[width=\linewidth]{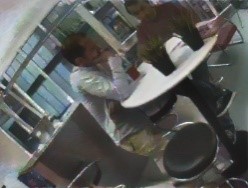} \\[2pt]
\rotatebox[origin=c]{90}{\parbox[c][1.8cm][c]{1.3cm}{\centering  \mbox{HyperE2VID}}} &
\includegraphics[width=\linewidth]{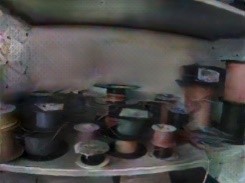} &
\includegraphics[width=\linewidth]{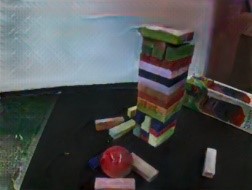} &
\includegraphics[width=\linewidth]{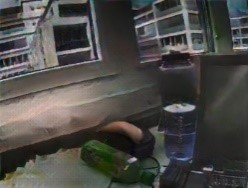} &
\includegraphics[width=\linewidth]{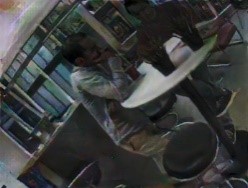} \\[2pt]
\rotatebox[origin=c]{90}{\parbox[c][1.8cm][c]{1.0cm}{\centering \mbox{ET-Net}}} &
\includegraphics[width=\linewidth]{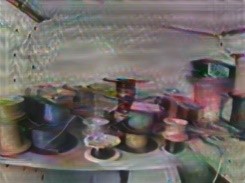} &
\includegraphics[width=\linewidth]{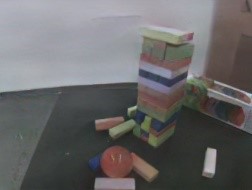} &
\includegraphics[width=\linewidth]{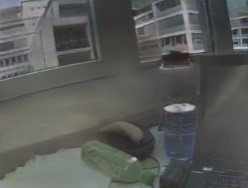} &
\includegraphics[width=\linewidth]{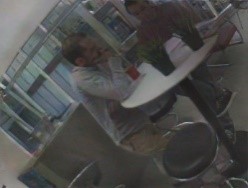} \\[2pt]
\rotatebox[origin=c]{90}{\parbox[c][1.8cm][c]{1.0cm}{\centering  E2VID+}} &
\includegraphics[width=\linewidth]{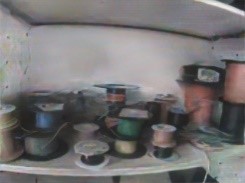} &
\includegraphics[width=\linewidth]{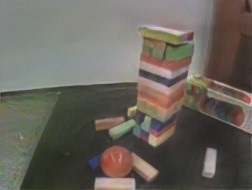} &
\includegraphics[width=\linewidth]{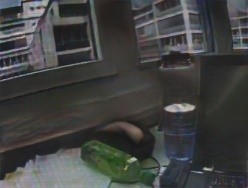} &
\includegraphics[width=\linewidth]{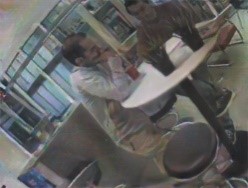} \\
\end{tabular}%
}
\caption{ Color reconstruction abilities of different methods. Overall, our model assigns vibrant and natural-looking colors with detailed scene reconstruction.}
\label{fig:color}
\vspace{-4mm}
\end{figure}

%% file: figure_robust.tex
\begin{figure*}[t]
    \centering
    \includegraphics[width=0.40\textwidth]{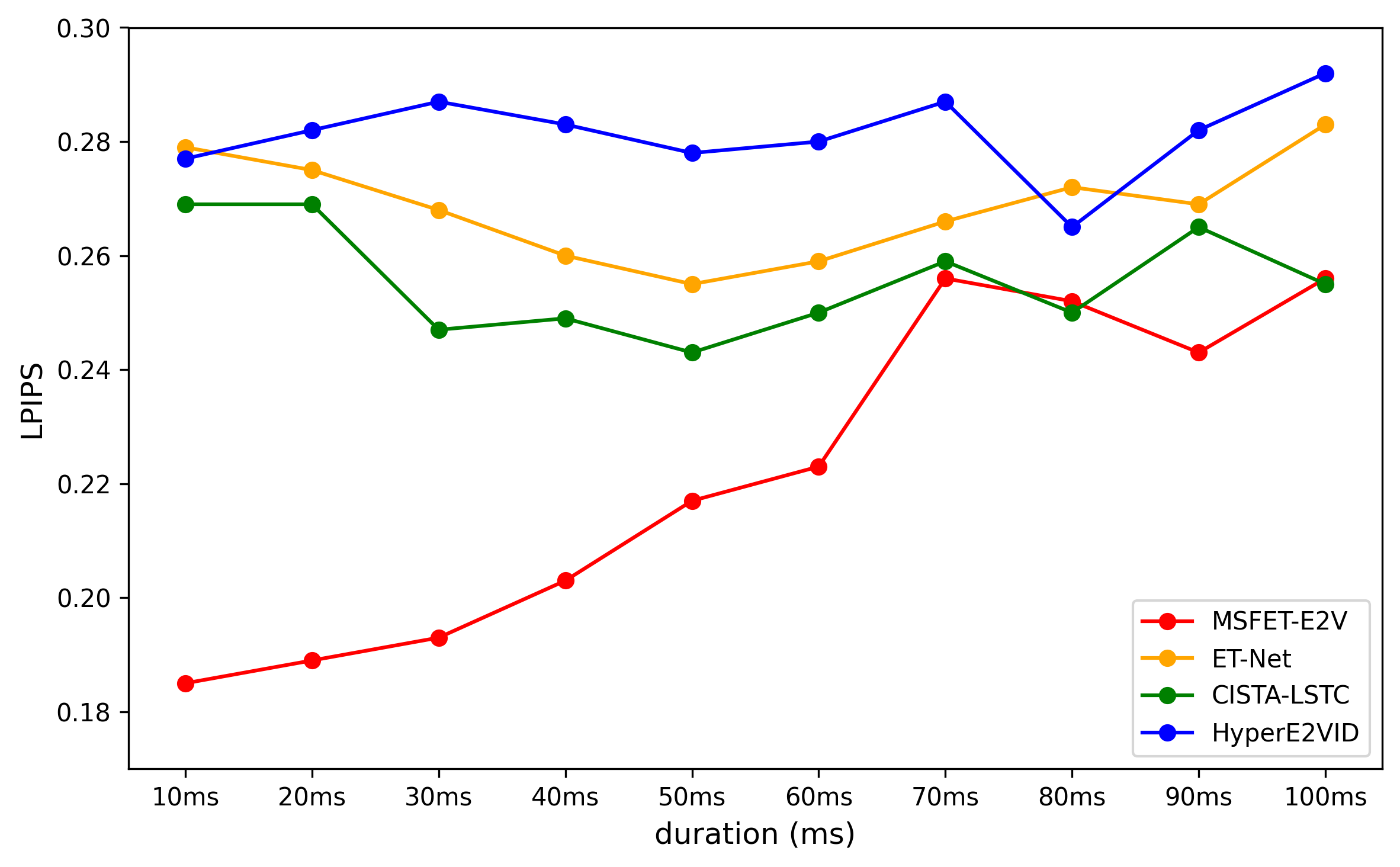}
    \includegraphics[width=0.40\textwidth]{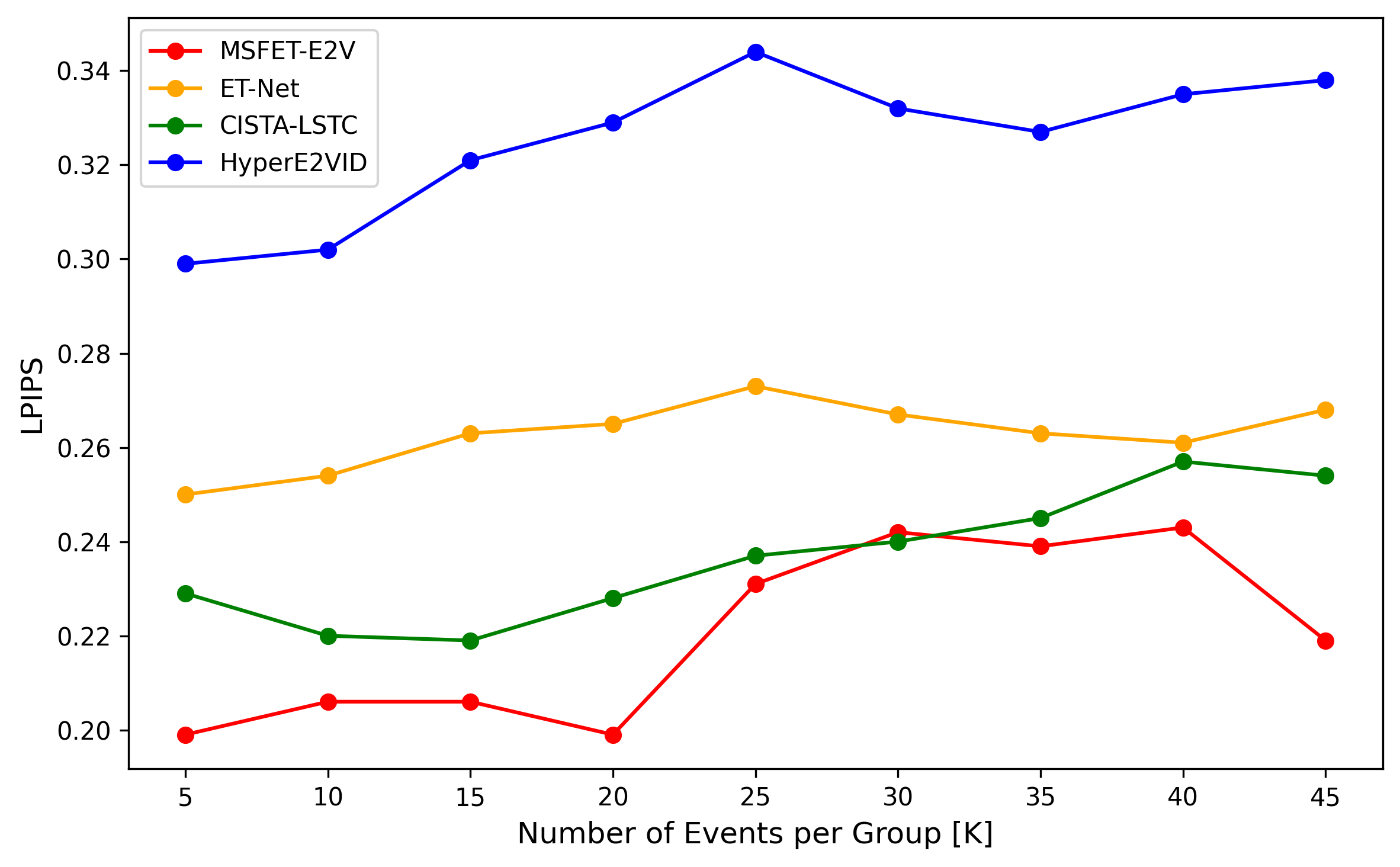}
    \caption{Comparison of LPIPS across different models. Left: different temporal windows grouping. Right: different number of events grouping.}
    \label{fig:robust}
\end{figure*}

%% file: tables_ablation.tex
\begin{table*}[t]
\centering
\caption{Ablation study on the components of the proposed model. We evaluate model variants on three event datasets. Here \textbf{bold} values indicate best-performing model.}
\vspace{-2mm}
\scriptsize
\setlength{\tabcolsep}{3pt}
\begin{tabular}{lccccc|ccc|ccc|ccc}
\toprule
\textbf{Model} & Spatio-temporal & Skip connection & Freq-domain & RB & WSB & \multicolumn{3}{c|}{\textbf{ECD}} & \multicolumn{3}{c|}{\textbf{HQF}} & \multicolumn{3}{c}{\textbf{MVSEC}} \\
\cmidrule(lr){7-9} \cmidrule(lr){10-12} \cmidrule(lr){13-15}
 & & & & & & PSNR ↑ & SSIM ↑ & LPIPS ↓ & PSNR ↑ & SSIM ↑ & LPIPS ↓ & PSNR ↑ & SSIM ↑ & LPIPS ↓ \\
\midrule
I & \checkmark & \checkmark & \xmark & \xmark & \xmark & 10.35 & 0.465 & 0.355 & 11.35 & 0.462 & 0.384 & 6.58 & 0.244 & 0.578 \\
II & \checkmark & \checkmark & \checkmark & \xmark & \xmark & 12.44 & 0.510 & 0.298 & 13.17 & 0.485 & 0.321 & 8.48 & 0.264 & 0.533 \\
III & \checkmark & \checkmark & \checkmark & \checkmark & \xmark & 13.27 & 0.534 & 0.286 & 14.23 & 0.517 & \textbf{0.289} & 10.05 & 0.297 & 0.521 \\
MSFET-E2V   & \checkmark & \xmark & \checkmark & \checkmark & \checkmark & \textbf{13.98} & \textbf{0.584} & \textbf{0.252} & \textbf{14.82} & \textbf{0.548} & 0.291 & \textbf{11.42} & \textbf{0.362} & \textbf{0.501} \\
\bottomrule
\end{tabular}
\label{Table:ablation_1}
\end{table*}
\begin{figure*}[ht]
\centering
\setlength{\tabcolsep}{0.5pt} 
\renewcommand{\arraystretch}{0.9} 
\begin{tabular}{@{} >{\centering\arraybackslash}m{0.16\textwidth} *{5}{>{\centering\arraybackslash}m{0.16\textwidth}} @{}}
\small Model I & \small Model II & \small Model III & \small MSFET-E2V & \small Event Voxel Grid & \small GT \\
\includegraphics[width=\linewidth,valign=m]{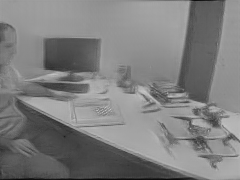} &
\includegraphics[width=\linewidth,valign=m]{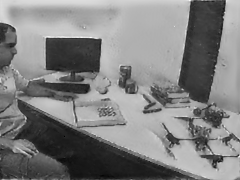} &
\includegraphics[width=\linewidth,valign=m]{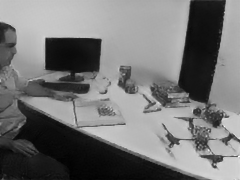} &
\includegraphics[width=\linewidth,valign=m]{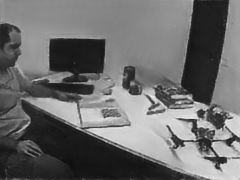} &
\includegraphics[width=\linewidth,valign=m]{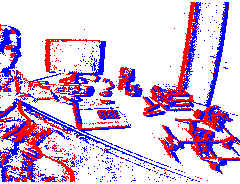} &
\includegraphics[width=\linewidth,valign=m]{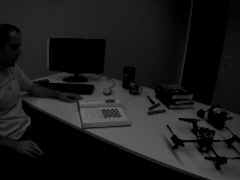} \\
0.52/0.36 & 0.41/0.43 & 0.39/0.46 & 0.32/0.50  \\
\end{tabular}
\caption{Visual comparison of different variants of the proposed model along with their corresponding LPIPS/SSIM scores is presented. The best-performing model is highlighted in \textbf{bold}. Our proposed model, MSFET-E2V, demonstrates superior reconstruction quality and is particularly effective at preserving fine details.}
\label{fig:modules}
\end{figure*}

%% file: table_ablation_II.tex
\begin{table*}[t]
\centering
\caption{
Ablation study on the role of frequency-domain subbands in the CDAM. Performance comparison across three datasets using CDAM variants: CDAM\_LL (only LF subbands), CDAM\_HF (only HF subbands), and the complete model MSFET-E2V (LF + HF). Here \textbf{bold} values indicate best-performing model.}
\vspace{-2mm}
\begin{tabular*}{\textwidth}{@{\extracolsep{\fill}}lccc|ccc|ccc}
\toprule
\textbf{Datasets} & \multicolumn{3}{c|}{\textbf{ECD}} & \multicolumn{3}{c|}{\textbf{HQF}} & \multicolumn{3}{c}{\textbf{MVSEC}} \\
\cmidrule(lr){2-4} \cmidrule(lr){5-7} \cmidrule(lr){8-10}
\textbf{Model} & CDAM\_LL & CDAM\_HF & MSFET-E2V   & CDAM\_LL & CDAM\_HF & MSFET-E2V   & CDAM\_LL & CDAM\_HF & MSFET-E2V   \\
\midrule
\textbf{SSIM} $\uparrow$   & 0.499 & 0.543 & \textbf{0.584} & 0.462 & 0.485 & \textbf{0.548} & 0.299 & 0.324 & \textbf{0.362} \\
\textbf{LPIPS} $\downarrow$ & 0.322 & 0.283 & \textbf{0.252} & 0.344 & \textbf{0.289} & 0.291 & 0.578 & 0.533 & \textbf{0.502} \\
\textbf{PSNR} $\uparrow$   & 10.35 & 12.44 & \textbf{13.98} & 11.35 & 13.17 & \textbf{14.82} & 7.58 & 9.48 & \textbf{11.42} \\
\bottomrule
\end{tabular*}
\label{Table:ablation_cdam}
\end{table*}
\begin{figure*}[ht]
\centering
\setlength{\tabcolsep}{0.5pt} 
\renewcommand{\arraystretch}{1} 
\begin{tabular}{@{} >{\centering\arraybackslash}m{0.15\textwidth} *{5}{>{\centering\arraybackslash}m{0.15\textwidth}} @{}}
\small CDAM\_LL & \small CDAM\_HF & \small MSFET-E2V & \small Event Voxel Grid & \small GT \\
\includegraphics[width=\linewidth,valign=m]{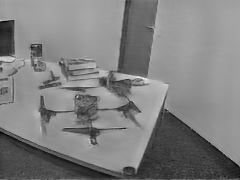} &
\includegraphics[width=\linewidth,valign=m]{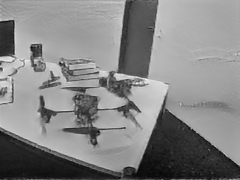} &
\includegraphics[width=\linewidth,valign=m]{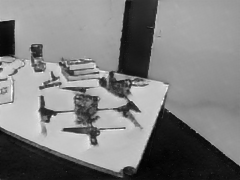} &
\includegraphics[width=\linewidth,valign=m]{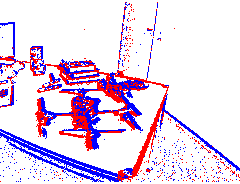} &
\includegraphics[width=\linewidth,valign=m]{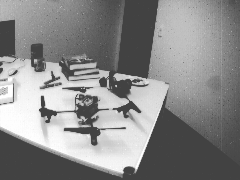} \\[-1pt]
0.33/0.47 & 0.32/0.49 & 0.25/0.52\\[-1pt]
\includegraphics[width=\linewidth,valign=m]{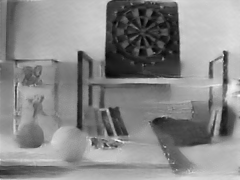} &
\includegraphics[width=\linewidth,valign=m]{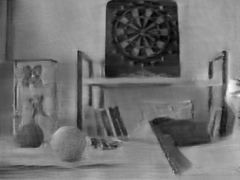} &
\includegraphics[width=\linewidth,valign=m]{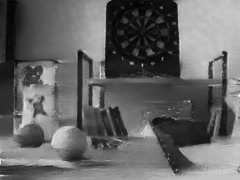} &
\includegraphics[width=\linewidth,valign=m]{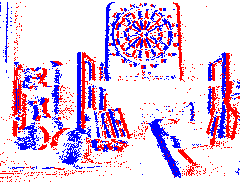} &
\includegraphics[width=\linewidth,valign=m]{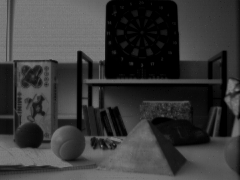}\\[-1pt]
0.30/0.46 & 0.28/0.47 & 0.25/0.49\\[-1pt]
\includegraphics[width=\linewidth,valign=m]{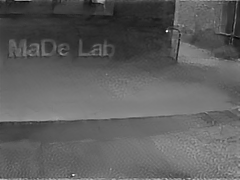} &
\includegraphics[width=\linewidth,valign=m]{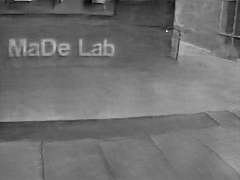} &
\includegraphics[width=\linewidth,valign=m]{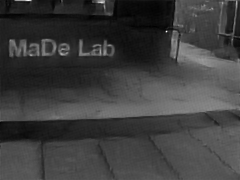} &
\includegraphics[width=\linewidth,valign=m]{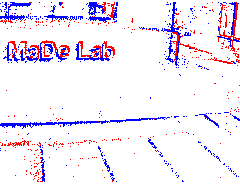} &
\includegraphics[width=\linewidth,valign=m]{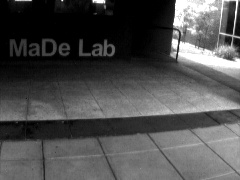} \\ [-1pt]
0.38/0.48 & 0.33/0.48 & 0.28/0.51\\[-1pt]
\includegraphics[width=\linewidth,valign=m]{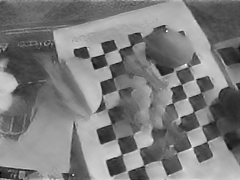} &
\includegraphics[width=\linewidth,valign=m]{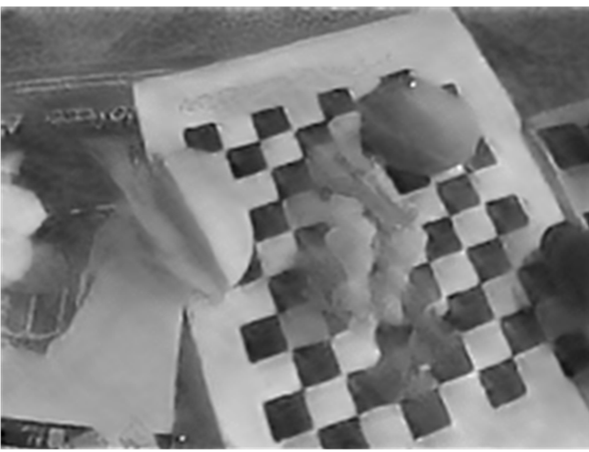} &
\includegraphics[width=\linewidth,valign=m]{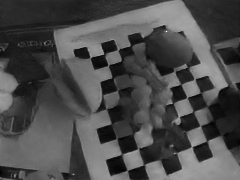} &
\includegraphics[width=\linewidth,valign=m]{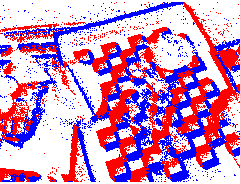} &
\includegraphics[width=\linewidth,valign=m]{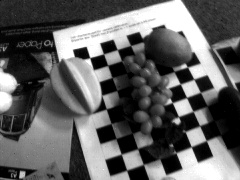} \\ [-1pt]
0.26/0.48 & 0.22/0.50 & 0.20/0.56\\[-1pt]
\end{tabular}
\caption{Visual Results on variants of CDAM. The reconstructed images are provided with LPIPS/SSIM score. Here \textbf{bold} values indicate best-performing model.}
\label{fig:ablation_cdam}
\end{figure*}
\begin{table*}[t]
\centering
\caption{
Ablation study on the effect of encoder-decoder depth $(d)$. 
We evaluate performance across three datasets for depths $d \in \{1, 2, 3, 4\}$. Here \textbf{bold} values indicate best-performing model.}
\vspace{-2mm}
\begin{tabular}{lccc|ccc|ccc|ccc}
\toprule
\textbf{Dataset} & \multicolumn{3}{c|}{$\boldsymbol{d=1}$} & \multicolumn{3}{c|}{$\boldsymbol{d=2}$} & \multicolumn{3}{c|}{$\boldsymbol{d=3}$} & \multicolumn{3}{c}{$\boldsymbol{d=4}$} \\
\cmidrule(lr){2-4} \cmidrule(lr){5-7} \cmidrule(lr){8-10} \cmidrule(lr){11-13}
& PSNR $\uparrow$ & SSIM $\uparrow$ & LPIPS $\downarrow$ 
& PSNR $\uparrow$ & SSIM $\uparrow$ & LPIPS $\downarrow$ 
& PSNR $\uparrow$ & SSIM $\uparrow$ & LPIPS $\downarrow$ 
& PSNR $\uparrow$ & SSIM $\uparrow$ & LPIPS $\downarrow$ \\
\midrule
\textbf{ECD} 
& 8.19  & 0.399 & 0.302 
& 10.35 & 0.496 & 0.286 
& \textbf{13.98} & \textbf{0.584} & \textbf{0.252} 
& 10.25 & 0.581 & 0.256 \\
\textbf{HQF} 
& 10.56 & 0.362 & 0.396 
& 11.96 & 0.496 & 0.324 
& 14.82 & \textbf{0.548} & \textbf{0.291} 
& \textbf{15.32} & 0.526 & 0.298 \\
\textbf{MVSEC} 
& 7.86  & 0.218& 0.605 
& 9.86  & 0.299 & 0.558 
& \textbf{11.42} & \textbf{0.362} & \textbf{0.501} 
& 11.42 & 0.358 & 0.504 \\
\bottomrule
\end{tabular}
\label{Table:depth_ablation}
\end{table*}
\begin{table*}[ht]
\centering
\scriptsize
\caption{Ablation on the number of voxel bins \( b \). Best results for each dataset are shown in \textbf{bold}.}
\vspace{-2mm}
\setlength{\tabcolsep}{1pt}
\renewcommand{\arraystretch}{1.1}
\begin{tabular}{l|ccc|ccc|ccc|ccc|ccc|ccc}
\toprule
\textbf{Dataset} & \multicolumn{3}{c|}{$b=2$} & \multicolumn{3}{c|}{$b=4$} & \multicolumn{3}{c|}{{$b=5$}} & \multicolumn{3}{c|}{$b=6$} & \multicolumn{3}{c|}{$b=8$} & \multicolumn{3}{c}{$b=10$} \\
\cmidrule(lr){2-4} \cmidrule(lr){5-7} \cmidrule(lr){8-10} \cmidrule(lr){11-13} \cmidrule(lr){14-16} \cmidrule(lr){17-19}
& PSNR $\uparrow$ & SSIM $\uparrow$ & LPIPS $\downarrow$ & PSNR $\uparrow$ & SSIM $\uparrow$ & LPIPS $\downarrow$ & PSNR $\uparrow$ & SSIM $\uparrow$ & LPIPS $\downarrow$ & PSNR $\uparrow$ & SSIM $\uparrow$ & LPIPS $\downarrow$ & PSNR $\uparrow$ & SSIM $\uparrow$ & LPIPS $\downarrow$  & PSNR $\uparrow$ & SSIM $\uparrow$ & LPIPS $\downarrow$  \\
\midrule
\textbf{ECD }   & 12.92 & 0.522 & 0.347 & 13.98 & 0.579 & 0.264 & \textbf{13.98} & \textbf{0.584} & 0.252 & 13.98 & 0.584 & \textbf{0.249} & 13.98 & 0.579 & 0.256 & 13.45 & 0.572 & 0.259 \\
\textbf{HQF }   & 12.30 & 0.498 & 0.369 & 14.32 & 0.544 & 0.291 & \textbf{14.82} & 0.548 & \textbf{0.291} & 14.82 & \textbf{0.549} & 0.291 & 14.80 & 0.548 & 0.291 & 13.55 & 0.541 & 0.312 \\
\textbf{MVSEC}  & 9.81  & 0.298 & 0.572 & 11.36 & 0.355 & 0.504 & \textbf{11.42} & \textbf{0.362} & \textbf{0.501} & 11.42 & 0.362 & 0.501 & 11.42 & 0.362 & 0.509 & 11.25 & 0.342 & 0.512 \\
\bottomrule
\end{tabular}
\label{Table:voxel_bins}
\end{table*}

%% file: references.bib
@ARTICLE{obj_track,
  author={Liu, Bingde and Xu, Chang and Yang, Wen and Yu, Huai and Yu, Lei},
  journal={IEEE Trans. Instrum. Meas.}, 
  title={Motion robust high-speed light-weighted object detection with event camera}, 
  year={2023},
  volume={72},
  number={},
  pages={1-13}
}

@InProceedings{auto_driv,
author={Li, Jinghang and Liao, Bangyan and Lu, Xiuyuan and Liu, Peidong and Shen, Shaojie and Zhou, Yi},
title={Event-aided time-to-collision estimation for autonomous driving},
booktitle = {ECCV},
month = {Sep.},
year={2024},
pages={57-73}
}

@ARTICLE{robot,
  author={Mahlknecht, Florian and Gehrig, Daniel and Nash, Jeremy and Rockenbauer, Friedrich M. and Morrell, Benjamin and Delaune, Jeff and Scaramuzza, Davide},
  journal={IEEE Robot. Autom. Lett}, 
  title={Exploring event camera-based odometry for planetary robots}, 
  year={2022},
  volume={7},
  number={4},
  pages={8651-8658}
}

@InProceedings{aug_reality,
author={Chakravarthi, Bharatesh and Verma, Aayush Atul and Daniilidis, Kostas and Fermuller, Cornelia and Yang, Yezhou},
title={Recent event camera innovations: A survey},
booktitle={ECCV},
month = {Sep.},
year={2024},
pages={342-376}
}

@article{vgg,
  title={Very deep convolutional networks for large-scale image recognition},
  author={Simonyan, Karen and Zisserman, Andrew},
  journal={arXiv preprint arXiv:1409.1556},
  year={2014}
}

@ARTICLE{transformer_2,
  author={Gu, Daxin and Li, Jia and Zhu, Lin},
  journal={IEEE Signal Process. Lett}, 
  title={Learning adaptive parameter representation for event-based video reconstruction}, 
  year={2024},
  volume={31},
  number={},
  pages={1950-1954}}

@article{dwt_dblur,
title = {Image deblurring method based on self-attention and residual wavelet transform},
journal = {Expert Syst. Appl},
volume = {244},
pages = {123005},
year = {2024},
author = {Bing Zhang and Jing Sun and Fuming Sun and Fasheng Wang and Bing Zhu},
  
}

@inproceedings{hypernetworks,
  title={Hypernetworks},
  author={Ha, David and Dai, Andrew and Le, Quoc V},
  booktitle={ICLR},
  year={2017},
  month={Apr.}
}

@article{imagenet,
author = {Russakovsky, Olga and Deng, Jia and Su, Hao and Krause, Jonathan and Satheesh, Sanjeev and Ma, Sean and Huang, Zhiheng and Karpathy, Andrej and Khosla, Aditya and Bernstein, Michael and Berg, Alexander C. and Fei-Fei, Li},
title = {ImageNet large scale visual recognition challenge},
year = {2015},
volume = {115},
number = {3},
journal = {Int. J. Comput. Vision},
pages = {211–252}
}

@inproceedings{wave_hm,
title = {Wavelet transform image coding using human visual system},
author = {Lee, Imgeun and Kim, Jongsik and Kim, Yougkyu and Kim, Seongman and Park, Gooman and Park, Kyu},
booktitle={APCCAS}, 
year = {1995},
month = {Dec.},
pages = {619-623}
}

@inproceedings{scse,
author={Roy, Abhijit Guha and Navab, Nassir and Wachinger, Christian},
title={Concurrent spatial and channel `squeeze {\&} excitation' in fully convolutional networks},
booktitle = {MICCAI},
month={Sep.},
year={2018},
pages={421-429}
}

@ARTICLE{E2VID,
  author={Rebecq, Henri and Ranftl, René and Koltun, Vladlen and Scaramuzza, Davide},
  journal={IEEE Trans. Pattern Anal. Mach. Intell}, 
  title={High Speed and High Dynamic Range Video with an Event Camera}, 
  year={2021},
  volume={43},
  number={6},
  pages={1964-1980}
}

@ARTICLE{SPADE-E2VID,
  author={Cadena, Pablo Rodrigo Gantier and Qian, Yeqiang and Wang, Chunxiang and Yang, Ming},
  journal={IEEE Trans. Image Process}, 
  title={{SPADE-E2VID}: Spatially-Adaptive Denormalization for Event-Based Video Reconstruction}, 
  year={2021},
  volume={30},
  number={},
  pages={2488-2500}}

@ARTICLE{HyperE2VID,
  author={Ercan, Burak and Eker, Onur and Saglam, Canberk and Erdem, Aykut and Erdem, Erkut},
  journal={IEEE Trans. Image Process}, 
  title={{HyperE2VID}: Improving event-based video reconstruction via hypernetworks}, 
  year={2024},
  volume={33},
  number={},
  pages={1826-1837}
}

@inproceedings{3dTracking,
author = {Kim, Hanme and Leutenegger, Stefan and Davison, Andrew},
year = {2016},
month = {Oct.},
pages = {349-364},
title = {Real-time {3D} reconstruction and {6-DOF} tracking with an event camera},
booktitle = {ECCV},
}

@ARTICLE{WaveCNet,
  author={Li, Qiufu and Shen, Linlin and Guo, Sheng and Lai, Zhihui},
  journal={IEEE Trans. Image Process}, 
  title={{WaveCNet}: Wavelet Integrated {CNNs} to Suppress Aliasing Effect for Noise-Robust Image Classification}, 
  year={2021},
  volume={30},
  number={},
  pages={7074-7089}
}

@inproceedings{evreal,
  title={{EVREAL}: Towards a comprehensive benchmark and analysis suite for event-based video reconstruction},
  author={Ercan, Burak and Eker, Onur and Erdem, Aykut and Erdem, Erkut},
  booktitle={IEEE/CVF},
  pages={3943-3952},
  month={Jun.},
  year={2023}
}

@ARTICLE{spike_fft,
  author={Yu, Jiajie and Lu, Xing and Guo, Lijun and Wang, Chong and Li, Guoqi and Qian, Jiangbo},
  journal={IEEE Trans. Circuits Syst. Video Technol}, 
  title={Event-based Video Reconstruction via Spatial-Temporal Heterogeneous Spiking Neural Network}, 
  year={2025},
  volume={},
  number={},
  pages={1-1}

}

@InProceedings{trans_fft,
    author    = {Kim, Taewoo and Cho, Hoonhee and Yoon, Kuk-Jin},
    title     = {Frequency-aware Event-based Video Deblurring for Real-World Motion Blur},
    booktitle = {CVPR},
    month     = {Jun.},
    year      = {2024},
    pages     = {24966-24976}
}

@inproceedings{fet,
  title={Unlocking fine-grained details with wavelet-based high-frequency enhancement in transformers},
  author={Azad, Reza and Kazerouni, Amirhossein and Sulaiman, Alaa and Bozorgpour, Afshin and Aghdam, Ehsan Khodapanah and Jose, Abin and Merhof, Dorit},
  booktitle={MLMI},
  month={Oct.},
  pages={207-216},
  year={2023}
}

@InProceedings{wave-vit,
author={Yao, Ting and Pan, Yingwei and Li, Yehao and Ngo, Chong-Wah and Mei, Tao},
title={{Wave-ViT}: Unifying Wavelet and Transformers for Visual Representation Learning},
booktitle={ECCV},
year={2022},
month={Oct.},
pages={328-345}
}

@INPROCEEDINGS{FireNet,
  author={Scheerlinck, Cedric and Rebecq, Henri and Gehrig, Daniel and Barnes, Nick and Mahony, Robert E. and Scaramuzza, Davide},
  booktitle={WACV}, 
  title={Fast Image Reconstruction with an Event Camera}, 
  year={2020},
  month={Mar.},
  pages={156-163}
}

@INPROCEEDINGS{E2VID+,
author={Stoffregen, Timo
and Scheerlinck, Cedric
and Scaramuzza, Davide
and Drummond, Tom
and Barnes, Nick
and Kleeman, Lindsay
and Mahony, Robert},
booktitle={ECCV},
title={Reducing the Sim-to-Real Gap for Event Cameras},
year={2020},
month={Aug.},
pages={534-549}
}

@INPROCEEDINGS{ET-Net,
  author={Weng, Wenming and Zhang, Yueyi and Xiong, Zhiwei},
  booktitle={ ICCV}, 
  title={Event-based Video Reconstruction Using Transformer}, 
  year={2021},
  month={Oct.},
  pages={2543-2552}
}

@InProceedings{Temporal-Diff,
author={Zhu, Lin
and Zheng, Yunlong
and Zhang, Yijun
and Wang, Xiao
and Wang, Lizhi
and Huang, Hua},

title={Temporal Residual Guided Diffusion Framework for Event-Driven Video Reconstruction},
booktitle={ECCV},
year={2024},
month={Sep.},
pages={411-427}
}

@InProceedings{Day-Night,
author={Jeong, Yuhwan
and Cho, Hoonhee
and Yoon, Kuk-Jin},
title={Towards Robust Event-Based Networks for Nighttime via Unpaired Day-to-Night Event Translation},
booktitle={ECCV},
year={2024},
month={Sep.},
pages={286-306}
}

@ARTICLE{3D-DWT,
  title={Temporal-Aware Spiking Transformer Hashing Based on {3D-DWT}},
  author={Mei, Zihao and Li, Jianhao and Zhang, Bolin and Wang, Chong and Guo, Lijun and Li, Guoqi and Qian, Jiangbo},
  journal={arXiv preprint arXiv:2501.06786},
  year={2024}
}

@InProceedings{SWFormer,
author={Fang, Yuetong
and Wang, Ziqing
and Zhang, Lingfeng
and Cao, Jiahang
and Chen, Honglei
and Xu, Renjing},
title={Spiking Wavelet Transformer},
booktitle={ECCV},
year={2024},
month={Sep.},
pages={19-37}
}

@ARTICLE{Linear_basis,

AUTHOR={Ieng, Sio-Hoi and Lehtonen, Eero and Benosman, Ryad },

TITLE={Complexity Analysis of Iterative Basis Transformations Applied to Event-Based Signals},

JOURNAL={Front. Neurosci},

VOLUME={12},
pages={373},
YEAR={2018},
}

@InProceedings{unet,
author={Ronneberger, Olaf
and Fischer, Philipp
and Brox, Thomas},
title={{U-Net}: Convolutional Networks for Biomedical Image Segmentation},
booktitle={MICCAI},
month={Oct.},
year={2015},
pages={234-241}
}

@InProceedings{esim,
  title = 	 {{ESIM}: an Open Event Camera Simulator},
  author =       {Rebecq, Henri and Gehrig, Daniel and Scaramuzza, Davide},
  booktitle = 	 {CoRL},
  month={Oct.},
  pages = 	 {969-982},
  year = 	 {2018}
}

@InProceedings{mscoco,
author={Lin, Tsung-Yi
and Maire, Michael
and Belongie, Serge
and Hays, James
and Perona, Pietro
and Ramanan, Deva
and Doll{\'a}r, Piotr
and Zitnick, C. Lawrence},
title={Microsoft {COCO}: Common Objects in Context},
booktitle={ECCV},
year={2014},
month={Sep.},
pages={740-755}
}

@article{haar,
title = {The {Haar} wavelet transform: its status and achievements},
journal = {Comput. Electr. Eng},
volume = {29},
number = {1},
pages = {25-44},
year = {2003},
author = {Radomir S. Stanković and Bogdan J. Falkowski}
}

@ARTICLE{ELWNet,
  author={Wang, Zhenyu and Zhang, Yunzhou and Liu, Yan and Zhu, Delong and Coleman, Sonya A. and Kerr, Dermot},
  journal={IEEE Trans. Circuits Syst. Video Technol}, 
  title={{ELWNet}: An Extremely Lightweight Approach for Real-Time Salient Object Detection}, 
  year={2023},
  volume={33},
  number={11},
  pages={6404-6417}
}

@ARTICLE{Davis240C,
  author={Brandli, Christian and Berner, Raphael and Yang, Minhao and Liu, Shih-Chii and Delbruck, Tobi},
  journal={ IEEE J. Solid-State Circuits}, 
  title={A 240 × 180 130 dB 3 µs Latency Global Shutter Spatiotemporal Vision Sensor}, 
  year={2014},
  volume={49},
  number={10},
  pages={2333-2341}
  }

@inproceedings{WaveConv,
author = {Finder, Shahaf E. and Amoyal, Roy and Treister, Eran and Freifeld, Oren},
title = {Wavelet Convolutions for Large Receptive Fields},
year = {2024},
pages = {363-380},
booktitle = {ECCV },
month={Sep.}
}

@inproceedings{convlstm,
author = {Shi, Xingjian and Chen, Zhourong and Wang, Hao and Yeung, Dit-Yan and Wong, Wai-Kin and Woo, Wang-Chun},
title = {Convolutional {LSTM} Network: A Machine Learning Approach for Precipitation Nowcasting},
booktitle = {NeurIPS},
month={Dec.},
pages = {802-810},
year = {2015}
}

@InProceedings{voxel,
author={Zhu, Alex Zihao
and Yuan, Liangzhe
and Chaney, Kenneth
and Daniilidis, Kostas},
title={Unsupervised Event-Based Optical Flow Using Motion Compensation},
booktitle="ECCV",
year="2019",
month={Sep.},
pages={711-714}
}

@InProceedings{ced,
  title={{CED}: Color event camera dataset},
  author={Scheerlinck, Cedric and Rebecq, Henri and Stoffregen, Timo and Barnes, Nick and Mahony, Robert and Scaramuzza, Davide},
  booktitle={CVPRW},
month={Jun.},
  year={2019}
}

@article{ecd,
  title={The event-camera dataset and simulator: Event-based data for pose estimation, visual odometry, and {SLAM}},
  author={Elias Mueggler and Henri Rebecq and Guillermo Gallego and Tobi Delbr{\"u}ck and Davide Scaramuzza},
  journal={Int. J. Robot. Res},
  year={2016},
  volume={36},
  pages={142-149}
  
}

@ARTICLE{mvsec,
  author={Zhu, Alex Zihao and Thakur, Dinesh and Özaslan, Tolga and Pfrommer, Bernd and Kumar, Vijay and Daniilidis, Kostas},
  journal={IEEE Robot. Autom. Lett}, 
  title={The Multivehicle Stereo Event Camera Dataset: An Event Camera Dataset for {3D} Perception}, 
  year={2018},
  volume={3},
  number={3},
  pages={2032-2039}

}

@ARTICLE{deep_unfold_TIP,
  author={Ge, Chengjie and Fu, Xueyang and Wang, Kunyu and Zha, Zheng-Jun},
  journal={IEEE Trans. Image Process}, 
  title={Event-Based Video Reconstruction With Deep Spatial-Frequency Unfolding Network}, 
  year={2025},
  volume={34},
  number={},
  pages={1779-1794}

}

@ARTICLE{cista-lstc,
  author={Liu, Siying and Dragotti, Pier Luigi},
  journal={IEEE Trans. Pattern Anal. Mach. Intell}, 
  title={Sensing Diversity and Sparsity Models for Event Generation and Video Reconstruction from Events}, 
  year={2023},
  volume={45},
  number={10},
  pages={12444-12458}
}
